\documentclass[runningheads]{llncs}
\usepackage{graphicx}
\usepackage{comment}
\usepackage{amsmath,amssymb} % define this before the line numbering.
\usepackage{color}
\usepackage{xcolor}
\usepackage{booktabs}
\usepackage{comment}
\usepackage{graphicx}
\usepackage{gensymb}
\usepackage{multirow}
\usepackage{subfig}
\usepackage[resetlabels]{multibib}
\newcites{A}{References}

\newcommand\scalemath[2]{\scalebox{#1}{\mbox{\ensuremath{\displaystyle #2}}}}

\usepackage{mathtools}

\usepackage{tabularx}
\usepackage[export]{adjustbox}% <-- added
\usepackage{floatrow}
\newfloatcommand{capbtabbox}{table}[][\FBwidth]

\definecolor{gucolor}{RGB}{50,200,0}

\newcommand\Set[2]{\{\,#1\mid#2\,\}}

\DeclareMathOperator{\avg}{avg}
\DeclareMathOperator{\pself}{Self6D}
\DeclareMathOperator{\pselfLB}{Self6D-LB}
\DeclareMathOperator{\pselfUB}{Self6D-UB}

\newcommand{\etal}[0]{\textit{et al.}}
\newcommand{\ie}[0]{\textit{i.e.}}
\newcommand{\wrt}[0]{\textit{w.r.t.~}}
\newcommand{\eg}[0]{\textit{e.g.}}
\newcommand{\vs}[0]{\textit{v.s.~}}
\newcommand{\loss}[0]{\mathcal{L}}
\usepackage{hyperref}

\newcommand*\samethanks[1][\value{footnote}]{\footnotemark[#1]}

\usepackage[width=122mm,left=12mm,paperwidth=146mm,height=193mm,top=12mm,paperheight=217mm]{geometry}

\begin{document}

\title{Self6D: Self-Supervised Monocular 6D Object Pose Estimation}

\titlerunning{Self6D: Self-Supervised Monocular 6D Object Pose Estimation}

\author{Gu Wang$^{1,2,}$\thanks{equal contribution} \quad Fabian Manhardt$^{2,}$\samethanks \quad Jianzhun Shao$^1$ \\ Xiangyang Ji$^1$ \quad Nassir Navab$^2$ \quad Federico Tombari$^{2,3}$}
\institute{$^1$Tsinghua University \quad $^2$Technical University of Munich \quad $^3$Google\\
{\tt\small \{wangg16, sjz18\}@mails.tsinghua.edu.cn, xyji@tsinghua.edu.cn, \\ 
\{fabian.manhardt, nassir.navab\}@tum.de, tombari@in.tum.de}}
\authorrunning{G. Wang \etal}

\maketitle

\begin{abstract} % -----------------------------------------------
6D object pose estimation is a fundamental problem in computer vision. Convolutional Neural Networks (CNNs) have recently proven to be capable of predicting reliable 6D pose estimates even from monocular images. Nonetheless, CNNs are identified as being extremely data-driven, and acquiring adequate annotations is oftentimes very time-consuming and labor intensive. To overcome this shortcoming, we propose the idea of monocular 6D pose estimation by means of self-supervised learning, 
removing the need for real annotations. After training our proposed network fully supervised with synthetic RGB data, we leverage recent advances in neural rendering to further self-supervise the model on unannotated real RGB-D data, seeking for a visually and geometrically optimal alignment. Extensive evaluations demonstrate that our proposed self-supervision is able to significantly enhance the model's original performance, outperforming all other methods relying on synthetic data or employing elaborate techniques from the domain adaptation realm.
\end{abstract}

\begin{figure}[t]
	\centering
	\includegraphics[width = 0.99\linewidth]{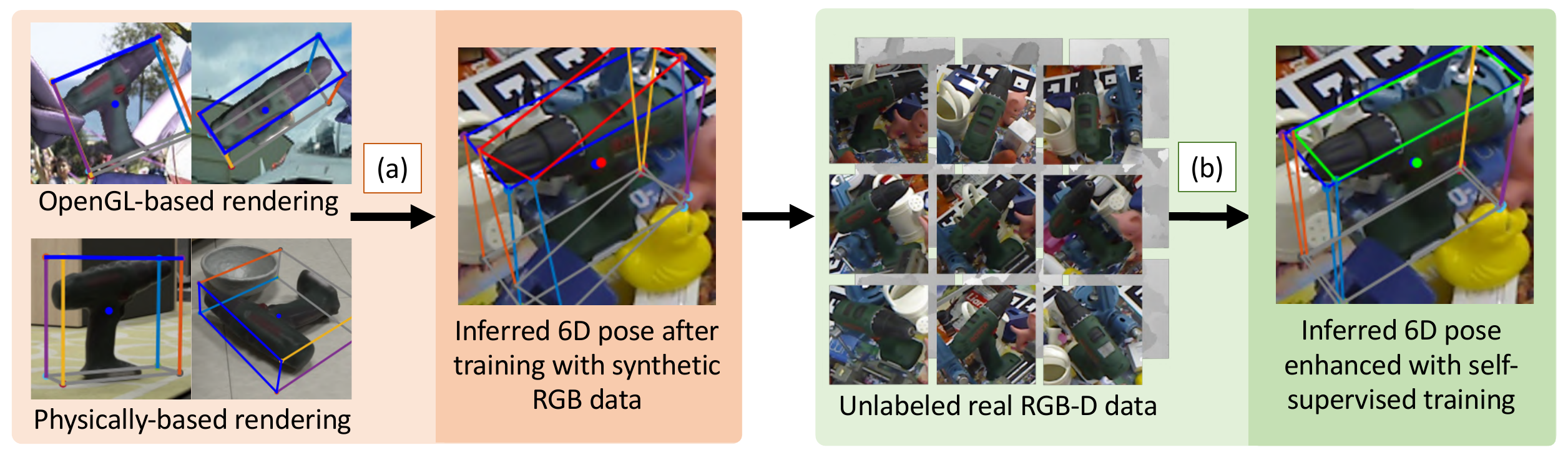}
% 	\vspace{-0.2mm}
	\caption{\textbf{Abstract illustration} of our proposed method. 
	We visualize the 6D pose by overlaying the image with the corresponding transformed 3D bounding box.
	To circumvent the use of real 6D pose annotations, we firstly train our model purely on synthetic RGB data (\textit{a}). 
	Secondly, employing a large amount of unlabeled real RGB-D images (\textit{b}), we significantly improve its performance (\textit{right}). 
	While \textit{Blue} constitutes the ground truth pose, we demonstrate in \textit{Red} and \textit{Green} the results before and after applying our self-supervision, respectively.
	}
% 	\vspace{-2mm}
	\label{fig:intro}
\end{figure}

\section{Introduction}

While learning-based techniques have recently demonstrated great performance in estimating the 6D pose (\ie~the 3D translation and rotation), a huge amount of training data is required~\cite{li2019cdpn,peng2019pvnet,tekin18_yolo6d}.
Furthermore, contrary to most 2D computer vision tasks such as classification, object detection and segmentation, acquiring real world 6D object pose annotations is much more labor intensive, time consuming, and error-prone~\cite{hodan2017t,kaskman2019homebreweddb}.

In order to deal with the lack of real annotations, one common approach is to simulate a large amount of synthetic images~\cite{richter2016playing,Su_2015_ICCV}. This is especially appealing for object pose estimation as one usually aims at estimating the 6D pose from an image \wrt the corresponding CAD model. Knowing the CAD model enables easy generation of enormous RGB images by randomly sampling 6D poses. Many approaches typically rely on rendering the models using OpenGL and placing them on random background images (drawn from large-scale 2D object datasets such as COCO~\cite{coco_eccv14}) in order to impose invariance to changing scenes~\cite{kehl2017ssd,manhardt2018deep}. 
Recent works propose to instead employ physically-based rendering to produce high quality renderings, and additionally enforce real physical constraints, as they can provide additional cues for the 6D pose~\cite{hodan2019photorealistic,tremblay2018deep}. 

Despite compelling results, these methods usually still exhibit inferior performance when inferring from real world data, due to the withstanding domain gap between real and synthetic data. 
Although techniques for domain adaption~\cite{bousmalis2017unsupervisedPixelda}, domain randomization~\cite{sundermeyer2018implicit} and photorealistic rendering~\cite{hodan2019photorealistic} can mitigate the problem to some extent, the performance is still far from satisfactory. 

This motivated us to investigate the problem from an entirely different angle. 
Humans have the amazing ability to learn about the 3D world, whilst only perceiving it through 2D images. 
Moreover, they can even learn 3D world properties without supervision from another human or \textit{labels} in a self-supervised fashion through making observations and validating if these observations are in accordance with the expected outcome~\cite{spelke1990principles}. 
In our context, while labeling the 6D pose is a severe bottleneck, recording unannotated data can be easily achieved at scale. 
Therefore, similar to learning for humans, we aim at teaching a neural network to reason about the 6D pose of an object by leveraging these unsupervised examples. 
As shown in Fig.~\ref{fig:intro}, we first train our method fully-supervised with synthetic data. Afterwards, employing unannotaed RGB-D data, we make use of self-supervised learning to enhance the model's performance on real data. 

To accomplish this, it is required to understand 3D properties solely from 2D images. 
The mechanism of experiencing the 3D world as images on the eye's retina is known as \textit{rendering} and has been also extensively explored in Computer Graphics~\cite{marschner2015CG}.  Unfortunately, rendering is also known to be non-differentiable due to the rasterization step, as gradients cannot be computed for the $argmax$ function. Nevertheless, many approaches for differentiable rendering have been recently proposed. The real gradient is thereby either approximated~\cite{kato2018renderer,opendr_eccv14}, or computed analytically by approximating the rasterization function itself~\cite{liu2019softras,chen2019learning_dibrenderer}.

In summary, we make the following contributions. i) To the best of our knowledge, we are the first to conduct self-supervised 6D object pose estimation from real data, without the need of 6D labels. ii) Leveraging neural rendering, we formulate a self-supervised 6D pose estimation solution by means of visual and geometric alignment. iii) We experimentally show that the proposed method, which we dub $\pself$, outperforms state-of-the-art methods for monocular 6D object pose estimation trained without real annotations by a large margin.
\section{Related work}
We first introduce recent work in monocular 6D pose estimation. Afterwards, we discuss important methods from neural rendering as they form a core part of our (as well as other) self-supervised learning frameworks. We then outline other successful approaches grounded on self-supervised learning. Lastly, we take a brief look at domain adaptation in the field of 6D pose, since our method can be considered an implicit formulation to close the synthetic-to-real domain gap.

\subsection{Monocular 6D Pose Estimation}
Recently, monocular 6D pose estimation has received a lot of attention and several %new 
very promising works have been proposed~\cite{hodan2018bop}.

One major branch is grounded on establishing 2D-3D correspondences between the image and the 3D CAD model. After estimating these correspondences, P$n$P is commonly employed to solve for the 6D pose. Inspired by~\cite{Brachmann2014Learning6O,brachmann2016uncertainty}, Rad~\etal~propose to employ a CNN to estimate the 2D projections of the 3D bounding box corners in image space~\cite{rad2017bb8}. Similarly, \cite{hu2019segpose,peng2019pvnet} also regress 2D projections of associated sparse 3D keypoints, however, both employ segmentation paired with voting to improve the reliability. In contrast, \cite{zakharov2019dpod,li2019cdpn,park2019pix2pose} ascertain dense 2D-3D correspondences, rather than sparse ones. % While \cite{li2019cdpn} decouples the estimation of translation and rotation, \cite{park2019pix2pose} leverages recent advances in GANs to further strengthen the robustness towards occlusion.

Another branch of work learns a pose embedding, which can be utilized for latter retrieval. In particular, inspired by~\cite{wohlhart2015learning,kehl2016deep}, 
%Sundermeyer \etal employ 
\cite{sundermeyer2018implicit} employs an Augmented AutoEncoder (AAE) to learn latent representations for the 3D rotation.

A few methods also directly regress the 6D pose. For instance, while \cite{kehl2017ssd} extends \cite{liu2016ssd} to also classify the viewpoint and in-plane rotation, \cite{manhardt2019ambiguity} further adjusts \cite{kehl2017ssd} to implicitly deal with ambiguities %by means of
via multiple hypotheses (MHP). In \cite{xiang2017posecnn} and \cite{li2019deepim} the authors minimize a point matching loss.

The majority of these methods~\cite{hu2019segpose,park2019pix2pose,rad2017bb8,tekin18_yolo6d,xiang2017posecnn} exploit annotated real data to train their models. However, labeling real data commonly comes with a large cost in time and labor. Moreover, a shortage of sufficient real world annotations can lead to overfitting, regardless of exploiting strategies such as \emph{crop\&paste}~\cite{dwibedi2017cut,kaskman2019homebreweddb}. Other works, in contrast, fully rely on synthetic data to deal with these pitfalls~\cite{sundermeyer2018implicit,manhardt2019ambiguity}. Nonetheless, the performance falls far behind the methods based on real data. We, thus, harness the best of both worlds. While unannotated data can be easily obtained at scale, this combined with our self-supervision for pose is able to outperform all methods trained on synthetic data by a large margin.
%In addition, having enough data the method can generalize well, while still producing much more precise results than any method employing synthetic data only.

% ----------------------------------------------
\subsection{Neural Rendering}
Rasterization is a core part of all traditional rendering pipelines. Nonetheless, rasterization involves discrete assignment operations, preventing the flow of gradients throughout the rendering process. A series of work have been devoted to circumvent the hard assignment in order to reestablish the gradient flow.

Loper and Black introduce the first differentiable renderer by means of first-order Taylor approximation to calculate the derivative of pixel values~\cite{opendr_eccv14}. %with respect to the 2D pixel positions by image-space first-order Taylor approximation. %In OpenDR a vertex can only receive gradients from neighboring pixels of its mesh face edge within a close distance. 
In~\cite{kato2018renderer}, the authors instead approximate the gradient as the potential change of the pixel's intensity \wrt the meshes' vertices.
\emph{SoftRas}~\cite{liu2019softras} conducts rendering by aggregating the probabilistic contributions of each mesh triangle in relation to the rendered pixels. 
Consequently, the gradients can be calculated analytically, however, with the cost of extra computation. 
\emph{DIB-R}~\cite{chen2019learning_dibrenderer} further extends \cite{liu2019softras} to render of a variety of different lighting conditions.  %by considering rasterization as a combination of weighted interpolation of local mesh properties for foreground and global aggregation for background, enabling the rendering of a variety of different lighting conditions. 
In this work, we use \emph{DIB-R}~\cite{chen2019learning_dibrenderer} since it can be considered state-of-the-art for neural rendering.

%-----------------------------------------------
\subsection{Recent Trends in Self-Supervised Learning}
Self-supervised learning, \ie~learning despite the lack of properly labeled data, has recently enabled a large number of applications ranging from 2D image understanding all the way down to depth estimation for autonomous driving. 
In the core, self-supervised learning approaches implicitly learn about a specific task through solving related proxy tasks. 
This is commonly achieved by enforcing different constraints such as pixel consistencies across multiple views or modalities. 

One prominent approach in this area is \emph{MonoDepth}~\cite{godard2017unsupervised}, 
% proposed by Godard~\etal, 
which conducts monocular depth estimation
% In particular, the authors teach a neural network to estimate the depth of an image, 
by warping the 2D image points into another view and enforcing a minimum reprojection loss. 
In the following many works to extend \emph{MonoDepth} have been introduced~\cite{pillai2019superdepth,godard2019digging,Guizilini_2020_CVPR}. %Then a pixel consistency loss is enforced in the second view to supervise depth prediction. 
%For instance, \cite{guizilini2019packnet} employs velocity data from the car to learn depth estimation as well as ego-camera pose by warping the predictions forward and backward through time.
In visual representation learning, consistency is ensured by solving pretext tasks~\cite{kolesnikov2019revisiting}. %,zhang2016colorful,pathak2016context,noroozi2016unsupervised}. % including image colorization~\cite{zhang2016colorful}, image in-painting~\cite{pathak2016context}, `jigsaw' puzzles~\cite{noroozi2016unsupervised}, among others.
Another line of works explore self-supervised learning for 3D human pose estimation, leveraging multi-view epipolar geometry~\cite{kocabas2019self} or imposing 2D-3D consistency after lifting and reprojection of keypoints~\cite{chen2019unsupervised}. 
Self-supervised learning approaches using neural rendering have also been proposed in the field of 3D object and human body reconstruction from single RGB images~\cite{tung2017self,cmrKanazawa18,omran2018neural,alldieck2019learning,Zuffi_2019_ICCV}.

In the domain of 6D pose estimation, self-supervised learning is still a rather unexplored field. 
% In \cite{deng2020self}, the authors 
\cite{deng2020self} proposes a novel self-labeling pipeline with an interactive robotic manipulator. 
Essentially, running several methods for 6D pose estimation, they can reliably generate precise annotations. 
% By means of camera pose tracking, these labels are then propagated through time to achieve high scalability. 
Nonetheless, the final 6D pose estimation model is still trained fully-supervised using the acquired data.
In this work, we propose to instead directly employ self-supervision for 6D pose by enforcing visual and geometric consistencies on top of neural rendering.

% ------------------------------------
\subsection{Domain Adaptation for 6D Pose Estimation}
Bridging the domain gap between synthetic and real data is 
% a crucial task 
crucial in 6D pose estimation. 
% A lot of
Many works tackle this problem 
by learning a transformation to align the synthetic and real domains
% with the help of 
via Generative Adversarial Networks (GANs) ~\cite{bousmalis2017unsupervisedPixelda,lee2018diverse} or by means of feature mapping~\cite{rad2018domain}.
% The generator thereby learns a transformation which aligns the samples originating from the synthetic to the real domain. 
Exemplary, \cite{lee2018diverse} uses a cross-cycle consistency loss based on disentangled representations to embed images onto a domain-invariant content space and a domain-specific attribute space. \cite{rad2018domain} instead maps the features of a color-based pose estimator to a depth-based pose estimator.

In contrast, works from domain randomization aim at learning domain-invariant attributes. 
For instance, harnessing random backgrounds and severe augmentations~\cite{kehl2017ssd,sundermeyer2018implicit} or employing adversarial training to generate backgrounds and image augmentations~\cite{zakharov2019deceptionnet}.
% maximally fooling the pose estimation network.
% achieving invariance towards domain specific attributes. 
% While~\cite{kehl2017ssd} harnesses COCO images as background and add a lot of augmentations in order to become invariant to the domain, \cite{zakharov2019deceptionnet} employs adversarial training to generate backgrounds and image augmentations, maximally fooling the pose estimation network.

\begin{figure}[t!]
	\centering
	\includegraphics[width = 0.95\linewidth]{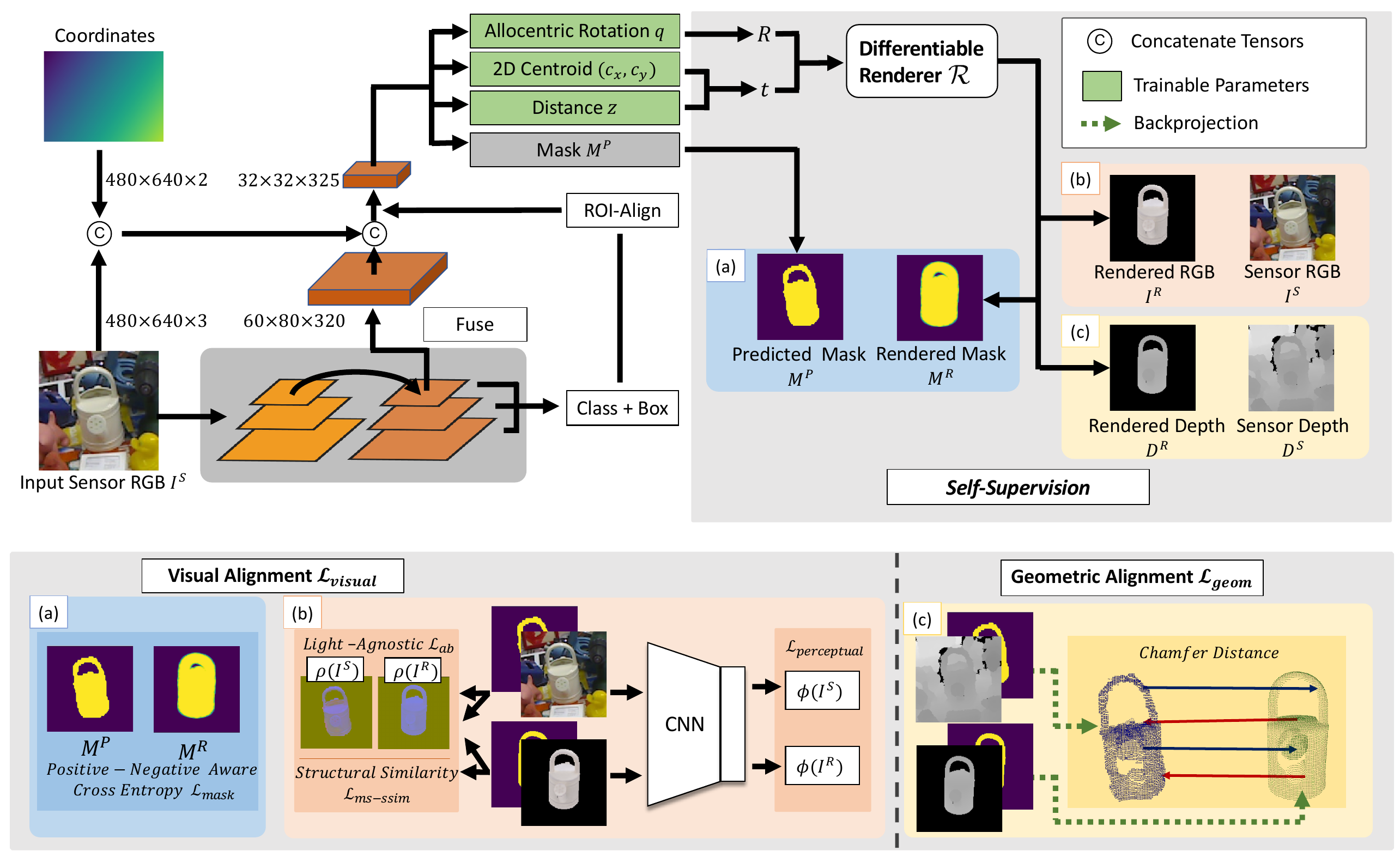}
% 	\vspace{-0.2mm}
	\caption{\textbf{Our self-supervised training pipeline.} 
	 \textit{Top}: We start training our model for 6D pose estimation purely on synthetic RGB data, to predict a 3D rotation $R$, translation $t$ and object instance mask $M^P$.
	 Using a large amount of unlabeled RGB-D images $(I^S, D^S)$, we enhance the model's performance by means of self-supervised learning. We 
% 	 use differentiable rendering %\emph{DIB-R} 
% 	 to
     differentiably
	 render ($\mathcal{R}$) the associated RGB-D image and mask $(I^R, D^R, M^R)$. 
	\textit{Bottom}: We impose various constraints to visually (\emph{a} and \emph{b}), and geometrically (\emph{c}) align the 6D pose.}
% 	\vspace{-2mm}
	\label{fig:arch}
\end{figure}
\section{Self-Supervised 6D Pose Estimation}
\label{sec:method}
In this work we aim at conducting 6D pose estimation from monocular images via %by means of 
self-supervised learning. To this end, we propose a novel model that can learn monocular pose estimation from both synthetic RGB data and real world unannotated RGB-D data.
Employing neural rendering, the model can be self-supervised by establishing coherence between real and rendered images \wrt the 6D pose. Since this requires good initial pose estimates, we rely on a two-stage approach. As shown in Fig.~\ref{fig:intro}, we start by training our model using synthetic RGB data only. Afterwards, we further enhance the pose estimation performance by leveraging unlabeled real world RGB-D data. 

We harness different visual and geometric constraints to seek the best alignment \wrt 6D pose. Unfortunately, while a 3D model contains information about the visible and invisible regions, the depth map only covers the visible surface. 
This complicates supervision since the invisible points would mistakenly contribute to the alignment. 
Therefore, we aim to extract only the model's visible surface given the current pose. This can be achieved in different ways: by culling the hidden points, or simply rendering the object in its current pose. Since we are required to render color for visual alignment, we resort to rendering depth for visible surface extraction, as it comes with no extra cost in computation.

We use the differentiable renderer \emph{DIB-R} proposed by \cite{chen2019learning_dibrenderer} to render 6D pose estimates from our model. Since \emph{DIB-R} is only able to render RGB images and object masks, we extend it to also provide the depth map fully differentiably. 
We additionally modify the camera projection to conduct a real perspective projection
\footnote{The code of our extended renderer is available at \url{https://github.com/THU-DA-6D-Pose-Group/Self6D-Diff-Renderer}}. 
Given the estimated 6D pose as 3D rotation $R$, 3D translation $t$, together with the 3D CAD model $\mathcal{M}$ and the camera intrinsics matrix $K$, we render the triplet $(I^R, D^R, M^R)$ consisting of the rendered RGB image $I^R$, the rendered depth map $D^R$ and the rendered mask $M^R$
\begin{equation}
 \mathcal{R}(R, t, K, \mathcal{M}) = (I^R, D^R, M^R).
\end{equation}

\paragraph{\textbf{Architecture Details.}}

Besides rendering, also the prediction of the 3D rotation and translation has to be differentiable in order to allow backpropagation. While methods based on establishing 2D-3D correspondences are currently dominating the field, it is infeasible to resort to them as gradients cannot be computed for P$n$P. To this end, we rely on a similar network architecture as ROI-10D~\cite{manhardt2019roi}, since they directly estimate rotation and translation. Unfortunately, the predicted poses from ROI-10D are not accurate enough to match the demands of our self-supervision, thus, we base our method on the more recent FCOS~\cite{tian2019fcos} detector. Moreover, a crucial part of our subsequent self-supervision requires object instance masks. Since no annotations are provided, we further extend ROI-10D to also estimate the visible object mask $M^P$ for each detection.

Our model is grounded on the object detector FCOS using a ResNet-50 based feature pyramid network (FPN)~\cite{lin2017feature} backbone to compute 2D region proposals. 
The FPN feature maps from different levels are then fused and concatenated with the input RGB image and 2D coordinates~\cite{liu2018intriguing}, from which the regions of interest are extracted via ROI-Align %~\cite{he2017mask} 
to predict masks and poses. Inspired by ROI-10D, we use different branches to predict the 3D rotation $R$ parameterized as a 4D quaternion $q$, the 3D translation $t$ defined as the 2D projection $(c_x,c_y)$ of the 3D object centroid and the distance $z$, and the visible object mask $M^P$. % Each branch consists of 4 convolutional layers with Group Normalization~\cite{wu2018group} and Leaky ReLU~\cite{xu2015empirical}.

To train the first-stage, we use focal loss~\cite{lin2017focal} for classification and GIoU loss~\cite{rezatofighi2019generalized} for bounding box regression. We rely on the binary cross entropy loss for mask prediction. 
As~\cite{li2019deepim}, we use the average of distinguishable model points metric as objective function for pose. The final loss can be summarized as 
\begin{equation}
\label{eq:loss_base}
\loss_{synthetic} \coloneqq 
    \lambda_{class} \loss_{focal} + 
    \lambda_{box} \loss_{giou} + 
    \lambda_{mask} \loss_{bce} + 
    \lambda_{pose} \loss_{pose}, \quad
\end{equation}
\begin{equation}
\label{eq:loss_pose}
 \text{with} \quad \loss_{pose} \coloneqq 
    %  \frac{1}{|\mathcal{M}|} \sum_{\mathbf{x} \in \mathcal{M}}
    \underset{x \in \mathcal{M}}{\avg}
    \|(R\mathbf{x} + t) - (\bar{R}\mathbf{x} + \bar{t})\|_1,
\end{equation}
where $\lambda_{class}, \lambda_{box}, \lambda_{mask}$ and $\lambda_{pose}$ denote the balance factors for each task, $\mathcal{M}$ denotes the 3D model, and $\left[R|t\right], \left[\bar{R}| \bar{t}\right]$ represent the predicted and ground truth poses, respectively. We kindly refer to the supplementary material for more details on the employed hyper-parameters.

For simplicity of the following, we define all foreground and background pixels as 
$N_+ \coloneqq \Set{(i,j)}{\forall M^P(i,j)=1}$
and
$N_- \coloneqq \Set{(i,j)}{\forall M^P(i,j)=0}$.
% \begin{equation}
%     N_+ \coloneqq \Set{(i,j)}{\forall M^P(i,j)=1} \quad \text{and} \quad N_- \coloneqq \Set{(i,j)}{\forall M^P(i,j)=0}.
% \end{equation} 
We further denote all pixels together as $N=N_+ \cup N_-$.

\paragraph{\textbf{Neural Rendering for Visual Alignment.}}
The most intuitive way is to simply align the rendered image $I^R$ with the sensor image $I^S$, deploying directly a loss on both samples. However, as the domain gap between $I^S$ and $I^R$ turns out to be very large, this does not work well in practice. In particular, lightning changes as well as reflection and bad reconstruction quality (especially in terms of color) oftentimes cause a high error despite having good pose estimates, eventually leading to divergence in the optimization. Hence, in an effort to keep the domain gap as small as possible, we impose multiple constraints measuring different domain-independent properties. In particular, we assess different visual similarities \wrt mask, color, image structure, and high-level content.

Since object masks are naturally domain agnostic, they can provide a particularly strong supervision. 
As our data is unannotated we refer to our predicted masks $M^P$ for a weak supervision. However, due to imperfect predicted masks, we utilize a modified cross-entropy loss~\cite{jiang2019integral}, which recalibrates the weights of positive and negative regions
\begin{equation}\label{eg:loss_mask}
\loss_{mask} \coloneqq 
- \frac{1}{|N_+|} \sum_{j\in N_+} M_j^P \log M_j^R 
- \frac{1}{|N_-|} \sum_{j\in N_-} \log (1 - M_j^R).
\end{equation}
Although masks are not suffering from the domain gap, they discard a lot of valuable information. In particular, color information is often the only guidance to disambiguate the 6D pose, especially for geometrically simple objects.

Since the domain shift is at least partially caused by light, we attempt to decouple light prior to measuring color similarity. % To this end, we convert the images to LAB space.
Let $\rho$ denote the transformation from RGB to LAB space, additionally discarding the light channel, we evaluate color coherence on the remaining two channels according to
\begin{equation}\label{eg:loss_AB}
\loss_{ab} \coloneqq 
\frac{1}{|N_+|} 
    \sum_{j \in N} 
    \|\rho(I^S)_j \cdot M_j^P - \rho(I^R)_j\|_1.
\end{equation}
 
We also avail various ideas from image reconstruction and domain translation, as they succumb the same dilemma.  
We assess the structural similarity (SSIM) in the RGB space and additionally follow the common practice to use a multi-scale variant, namely MS-SSIM~\cite{zhao2016loss}
\begin{equation}
\loss_{ms\text{-}ssim} \coloneqq 
     1 - ms\text{-}ssim(I^S \odot M^P, I^R, s).
\end{equation}
Thereby, $\odot$ denotes the element-wise multiplication and $s=5$ is the number of employed scales. For more details on MS-SSIM, we kindly refer the readers to the supplement and~\cite{zhao2016loss}.

Another common practice is to appraise the perceptual similarity~\cite{johnson2016perceptual,zhang2018unreasonable} in the feature space. To this end, a pretrained deep neural network as AlexNet~\cite{krizhevsky2012imagenet} is typically employed to ensure low- and high-level similarity. We apply the perceptual loss at different levels of the CNN. Specifically, we extract the feature maps of $L=5$ layers and normalize them along the channel dimension. Then we compute squared $L_2$ distances of the normalized feature maps $\hat{\phi}^l(\cdot)$ for each layer $l$.
We average the individual contributions spatially and sum across all layers~\cite{zhang2018unreasonable}
\begin{equation}
\loss_{perceptual} \coloneqq  
    \sum_{l=1}^L\frac{1}{|N^l|}
    \sum_{j \in N^l}
    \|\hat{\phi}_{j}^l(I^S \odot M^P) - \hat{\phi}_{j}^l(I^R)\|_2^2.
\end{equation}

The visual alignment is then composed as the weighted sum over all four terms
\begin{equation}
    \loss_{visual} \coloneqq 
     \loss_{mask} + \alpha \loss_{ab} +
    \beta \loss_{ms\text{-}ssim} + \gamma \loss_{perceptual},
\end{equation}
where $\alpha$, $\beta$ and $\gamma$ denote the balance factors for $\loss_{ab}$, $\loss_{ms\text{-}ssim}$, and $\loss_{perceptual}$, respectively. We refer to the supplement for more details on the  hyper-parameters.

\paragraph{\textbf{Neural Rendering for Geometric Alignment.}}
Since the depth map only provides information for the visible areas, aligning it with the transformed 3D Model similar to Eq.~\ref{eq:loss_pose} harms performance. Therefore, we exploit the rendered depth map to enable comparison of the visible areas only. 
Nevertheless, employing a loss directly on both depth maps leads to bad correspondences as the points where the masks are not intersecting cannot be matched. 

Hence, we operate on the visible surface in 3D to find the best geometric alignment. 
We first backproject $D^S$ and $D^R$ using the corresponding masks $M^P$ and $M^R$ to retrieve the visible pointclouds $\mathcal{P}^S$ and $\mathcal{P}^R$ in camera space with
\begin{equation}
     \pi^{-1}(D,M,K) = 
     \Set{K^{-1}  
    %  \begin{pmatrix} x_j  \\ y_j \\ 1 \end{pmatrix} 
    \begin{bmatrix} x_j  & y_j & 1 \end{bmatrix}^{T} 
     \cdot D_j} 
     {\forall j \in M>0},
\end{equation}
\begin{equation}
  \mathcal{P}^S \coloneqq  \pi^{-1}(D^S,M^P,K), \quad \quad \mathcal{P}^R \coloneqq \pi^{-1}(D^R, M^R, K).
\end{equation}
Thereby, $(x_j,y_j)$ denotes the 2D pixel location of $j$ in $M$.

Since it is infeasible to estimate direct 3D-3D correspondences between $\mathcal{P}^S$ and $\mathcal{P}^R$, we refer to the chamfer distance to seek the best alignment in 3D 
\begin{equation}
\scalemath{1}{
    \loss_{geom} \coloneqq \frac{1}{|\mathcal{P}^S|}\sum_{p^S \in \mathcal{P}^S}\min_{p^R \in \mathcal{P}^R}\|p^S - p^R\|_{2} +  \frac{1}{|\mathcal{P}^R|}\sum_{p^R \in \mathcal{P}^R}\min_{p^S \in \mathcal{P}^S}\|p^S - p^R\|_{2}.
}
\end{equation}

% \noindent
The overall self-supervision is 
% a simple combination of our loss terms for visual and geometric alignment,
$\loss_{Self} \coloneqq \loss_{visual} + \eta \loss_{geom}$,
% \begin{equation}
%     \loss_{Self} \coloneqq \loss_{visual} + \eta \loss_{geom},
% \end{equation}
with $\eta$ denoting the balance factor of $\loss_{geom}$. An overview is also presented in Fig.~\ref{fig:arch}. 
Noteworthy, while we require RGB-D data for self-supervision, we do not need any depth data during latter inference.

% In particular, the pose predictor still receives only color images to learn predicting the 6D pose.%, however, the error is computed employing depth as an extra modality.

\section{Evaluation}

In this section, we first introduce our experimental setup. Afterwards, we present the analysis on the quality of predicted masks and different ablations to illustrate 
the effectiveness of our proposed self-supervised loss.
% the existence of a strong correlation between our self-supervision and the actual pose errors, and to constitute the importance of each individual loss term. 
We conclude by comparing our method with other% current
state-of-the-art methods for 6D pose estimation and domain adaptation. For better understanding, in addition to the results of $\pself$, we also evaluate our method using synthetic data only and additionally employing real 6D pose labels. Since they can be considered the lower and upper bound of our method, we refer to them $\pselfLB$ and $\pselfUB$ in the following.

%\subsection{Experimental Setup}

%We implemented our method in PyTorch and ran all our experiments on a Nvidia TitanX GPU. In the first stage, we trained our model with a batchsize of 12 for 8 epochs. 
%During self-supervision, we trained the model for another 100 epochs with a batchsize of 3. We utilized RAdam~\cite{liu2019radam} combined with Lookahead~\cite{zhang2019lookahead}. %The initial learning rate was set to $10^{-4}$ and decayed after $72\%$ of training. %the training phase using a cosine schedule~\cite{loshchilov-ICLR17SGDR}. 

\paragraph{\textbf{Synthetic Training Data.}}
%Since we aim at avoiding the use of any annotated real data, we train our base model purely from synthetic data. 
~\cite{tremblay2018falling} and~\cite{hodan2019photorealistic} recently proposed to employ photorealistic and physically plausible renderings to improve 2D detection and 6D pose estimation, in contrast to simple OpenGL rendering~\cite{kehl2017ssd}. 
In our experiments it turns out that a mixture of both approaches, together with a lot of augmentations (\eg~random Gaussian noise, intensity jitter), leads to best results. 
% We assume that the variations in the high quality renderings are too low, to match the rather imperfect real world. Adding low quality OpenGL renderings, thus, implicitly enforces robustness to these variations.

\paragraph{\textbf{Datasets.}}
To evaluate our proposed method we leverage the commonly used \emph{LineMOD} dataset~\cite{Hinterstoisser2012}, which consists of 15 sequences, %each possessing $\approx$~1.2k images with clutter and mild occlusion. 
Only 13 of these provide water-tight CAD models and we, therefore, remove the other two sequences. In~\cite{Brachmann2014Learning6O}, the authors propose to sample $15\%$ of the real data for training to close the domain gap. We use the same split, however discarding the pose labels. 
As second dataset, we utilize the recent \emph{HomebrewedDB}~\cite{kaskman2019homebreweddb} dataset. 
However, we only employ the sequence which covers three objects from \emph{LineMOD}, to depict that we can even self-supervise the same model in a new environment. % \todo{Add YCB}

To also show generalization to other common datasets for 6D pose, we demonstrate the effectiveness of our self-supervision on 5 objects from \emph{YCB-Video}~\cite{xiang2017posecnn} in the supplementary material.
To compare with domain adaptation based methods, we refer to the usual \emph{Cropped LineMOD} dataset~\cite{wohlhart2015learning} including center-cropped $64\times64$ patches of 11 different small objects in cluttered scenes imaged in various of poses. %\emph{Cropped LineMOD} is grounded on \emph{LineMOD} and features $\approx$~110k rendered source images, $\approx$~10k real-world target images, $\approx$~1k real images for validation, and $\approx$~2.6k test images from the target domain.

\paragraph{\textbf{Metrics for 6D Pose.}}
We report our results \wrt the 
ADD
% \emph{Average Distance of Distinguishable Model Points} (ADD) 
metric~\cite{Hinterstoisser2012}, measuring whether the average deviation of the transformed model points is less than $10\%$ of the object's diameter.
For \emph{symmetric} objects (\eg, \emph{Eggbox} and \emph{Glue} in \emph{LineMOD}) we rely on the 
%\emph{Average Distance of Indistinguishable Model Points} (ADD-S)
ADD-S metric, which instead measures the error as the average distance to the \emph{closest} model point~\cite{Hinterstoisser2012,hodan2016evaluation}.
% Similar to state-of-the-art, we employ ADD-S for  \emph{Eggbox} and \emph{Glue} from \emph{LineMOD} and utilize ADD for all remaining objects.
\begin{equation}
    \textbf{ADD} = \underset{x \in \mathcal{M}}{\avg} \|(Rx + t) - (\bar{R}x + \bar{t})\|_2,
\end{equation}
\begin{equation}
    \textbf{ADD-S} = \underset{x_{2} \in \mathcal{M}}{\avg}
    \min_{x_{1} \in \mathcal{M}} \|(Rx_{1} + t) - (\bar{R}x_{2} + \bar{t})\|_2.
\end{equation}

\begin{figure}[t!]
	\centering
	\includegraphics[width = 0.82\linewidth]{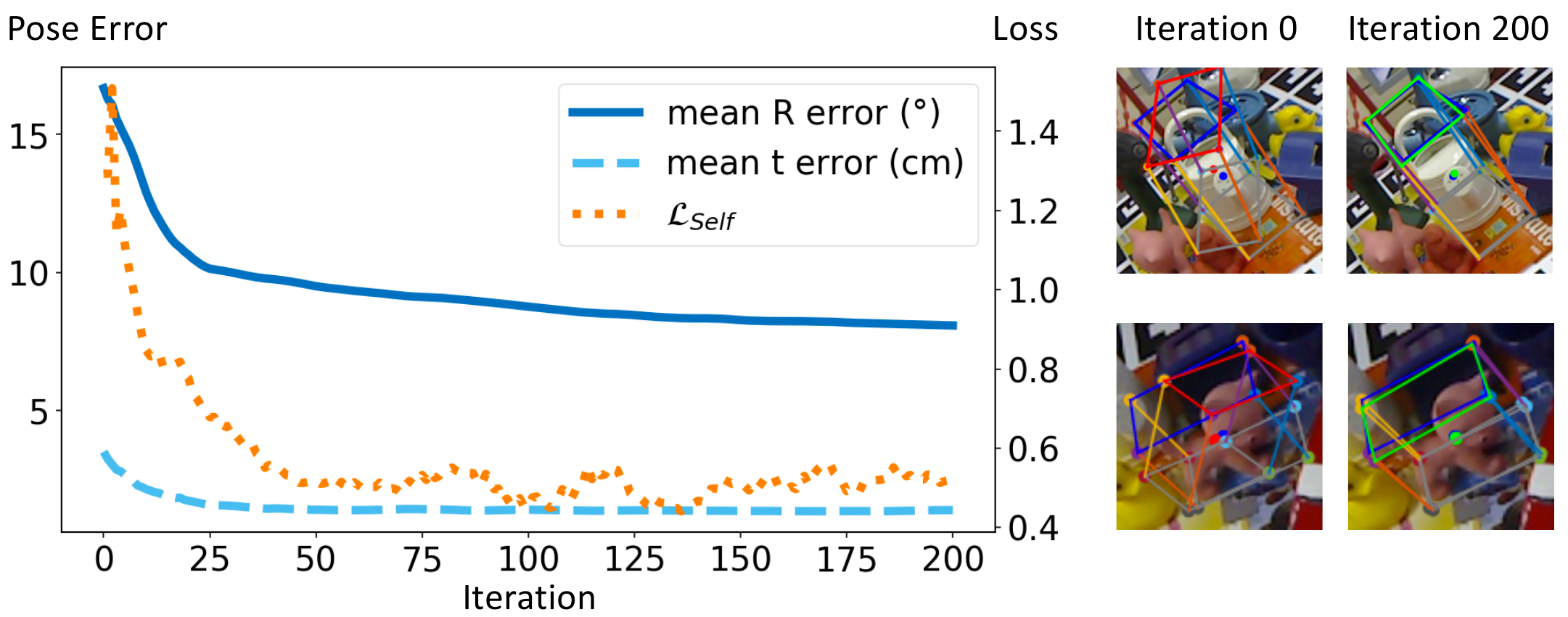}
	\caption{\textbf{Pose errors \vs self-supervision.} We optimize $\loss_{self}$ on single images from \emph{LineMOD} for 200 iterations and report the average over in total 100 images. We initialize the 6D poses with $\pselfLB$.}
	\label{fig:error_loss_vs_iter}
\end{figure}
\subsection{Analysis on the Quality of Predicted Masks}

Thanks to physically-based renderings, the predicted masks on the real data are very accurate, thus can be reliably used as a self-supervision signal. 
For instance, on the \emph{LineMOD} test set, the average F1 score and mIoU between the predicted masks and the ground-truth masks are 89.63\% and 90.38\%. Please refer to the supplementary for detailed results and qualitative examples. 

\subsection{Ablation Study}

\paragraph{\textbf{Self-Supervision \vs 6D Pose Error.}}

% Since we only implicitly solve for the 6D pose, w
We want to demonstrate that there is indeed a high correlation between our proposed $\loss_{Self}$ and the actual 6D pose errors. 
To this end, we randomly draw 100 samples from \emph{LineMOD} and optimize separately on each sample, always beginning from $\pselfLB$.
% the model previously trained on synthetic data.
Fig.~\ref{fig:error_loss_vs_iter} illustrates the average behavior \wrt loss \vs~6D pose error at each iteration. 
As the loss decreases, also the pose error for both, rotation and translation, continuously declines until convergence. The accompanying qualitative images (Fig.~\ref{fig:error_loss_vs_iter}, \textit{right}) further support this observation, as the initial pose is significantly worse compared to the final optimized result. We refer to the supplementary material for more qualitative results. 

\paragraph{\textbf{Individual Loss Contributions.}}
\begin{table}[t!]
\centering

\scalebox{0.8}{
\setlength{\tabcolsep}{0.2em}
\begin{tabular}{@{}l|c|c|c|c|c|c|c|c|c|c|c|c|c||c@{}}
\toprule
& Ape   & Bvise & Cam & Can & Cat & Drill & Duck  & Eggbox & Glue  & Holep & Iron  & Lamp  & Phone & Mean \\
\midrule
w/o $\loss_{mask}$ & 0.0 & 0.0 & 0.0 & 0.0 & 0.0 & 0.0 & 0.0 & 0.8 & 0.0 & 0.0 & 0.0 & 0.0 & 0.0 & 0.1 \\
w/o $\loss_{geom}$ & 0.0 & 10.1 & 3.1 & 0.0 & 0.0 & 7.5 & 0.1 & 33.0 & 0.2 & 0.0 & 5.9 & 20.7 & 2.4 & 6.4 \\
w/o $\loss_{ms\text{-}ssim}$ & 32.1 & 74.8 & 20.4 & 63.4 & 57.1 & 68.3 & 16.6 & \textbf{99.0} & 94.1 & 12.3 & 70.8 & \textbf{68.5} & 54.9 & 56.3 \\
w/o $\loss_{perceptual}$ & 34.9 & 74.4 & 33.5 & 64.8 & 55.3 & \textbf{70.0} & 17.2 & 98.7 & \textbf{94.8} & 10.7 & 76.3 & 68.1 & \textbf{56.5} & 58.1 \\
w/o $\loss_{ab}$ & \textbf{40.9} & 73.8 & 36.1 & 63.0 & \textbf{58.1} & 66.0 & 18.0 & 98.9 & 93.9 & \textbf{16.2} & 77.2 & 68.2 & 50.1 & 58.5 \\
$\pself$  & 38.9 & \textbf{75.2} & \textbf{36.9} & \textbf{65.6} & 57.9 & 67.0 & \textbf{19.6} & \textbf{99.0} & 94.1 & 15.5 & \textbf{77.9} & 68.2 & 50.1 & \textbf{58.9} \\
\midrule
$\pselfLB$ & 14.8 & 68.9 & 17.9 & 50.4 & 33.7 & 47.4 & 18.3 & 64.8  & 59.9 & 5.2 & 68.0 & 35.3 & 36.5 & 40.1 \\
$\pselfUB$ & 62.3 & 95.3 & 86.5 & 93.0 & 80.7 & 93.7   & 63.4 & 99.7 & 99.4 & 73.6 & 96.0 & 96.6 & 90.0 & 86.9 \\
\bottomrule
\end{tabular}
}
% \vspace{-0.2mm}
\caption{\textbf{Ablation.} We report the Average Recall of ADD(-S) on \emph{LineMOD}.}
% \vspace{-2mm}
\label{tab:lm_ablation}
\end{table}
% 
%Since our loss is a combination of multiple individual terms, 
% In Table~\ref{tab:lm_ablation} we illustrate
Table~\ref{tab:lm_ablation} illustrates
the contribution of each individual loss component on \emph{LineMOD}. 
% It turns out 
Note that supervision from both 
% domains, \ie~visual and geometry, 
visual and geometry domains
is vital for our self-supervised training.
% to enable training of our pose estimator. 
Disabling either $\loss_{mask}$ or $\loss_{geom}$ almost always leads to unstable training and divergence (the average recall is only $0.1\%$ and $6.4\%$ \wrt ADD(-S)). 
% Without these terms we are barely able to predict any correct pose and can, thus, only report an average recall of $0.1\%$ and $6.4\%$ \wrt ADD(-S). 
The remaining three factors, measuring color similarity, have a comparably small impact. %, yet, still further strengthen our models' capabilities. 
% In terms of numbers,
Concretely,
we drop by more than $2\%$ when disabling $\loss_{ms\text{-}ssim}$, and about $1\%$ referring to $\loss_{ab}$ and $\loss_{perceptual}$. Nonetheless, we still achieve the overall best results when applying all loss terms together. 
Most importantly, we can report a significant relative improvement of almost $50\%$ from $40.1\%$ to $58.9\%$ leveraging the proposed self-supervision. 
Moreover, except for the \emph{Duck} object, all other objects undergo a strong enhancement in ADD(-S). 
% In our experiments we 
% % discovered 
% found
% that a more diverse geometry and color simplifies latter supervision. However, since the \emph{Duck} is very limited in color and possesses a roundish shape, the total improvement only amounts to $1.4\%$. %This is expected given our proposed loss formulation. 
Noteworthy, we can almost halve the difference between training with and without real pose labels.

% \begin{comment}
% \paragraph{\textbf{Domain Adaption.}}
% \input{tables/table_hb}
% % 
% To better understand our abilities in the area of domain adaptation, in Table~\ref{tab:hb} we evaluate our method on three objects of \emph{HomebrewedDB}, which it shares with \emph{LineMOD}. Since our synthetic training data is generated using the camera intrinsics from the latter, we already have a large drop in performance. Moreover, when training on 15\% of \emph{LineMOD} the performance further decreases, since we adapt even more. Similarly, also our performance after self-supervising drops by a large margin, yet, considerably less. On the other hand, despite suffering from an original larger domain gap due to the wrong learnt intrinsics, we can still recover easily and adapt to the new camera and scene setup. This is also supported numerically as after self-supervision, we are almost on par with the corresponding results from Table~\ref{tab:lm_ablation}.
% \end{comment}

\subsection{Comparison with State-of-the-art}
In the first part of this section we present a comparison with current state-of-the-art methods in 6D pose estimation. In the latter part, we present our results in the area of domain adaptation referring to \emph{Cropped LineMOD}.

\subsubsection{6D Pose Estimation}

\paragraph{\textbf{LineMOD Dataset.}}
\begin{table}[t!]
\centering
\includegraphics[width = 0.965\linewidth]{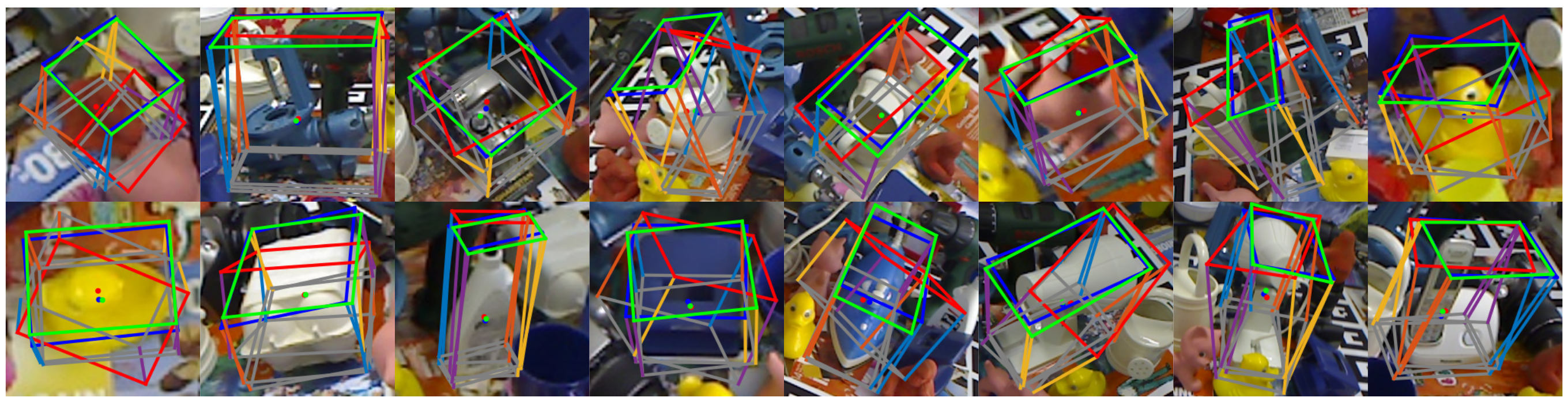}
\\
\scalebox{0.86}{
\setlength{\tabcolsep}{0.3em}
\begin{tabular}{@{}c|cccc|cccc@{}}
\toprule
Train data & \multicolumn{4}{c|}{ w/o Real Pose Labels} & \multicolumn{4}{c}{ with Real Pose Labels} \\
\midrule
Object & AAE\cite{sundermeyer2018implicit} & MHP\cite{manhardt2019ambiguity} & DPOD\cite{zakharov2019dpod} & $\pself$  & Tekin\cite{tekin18_yolo6d} & DPOD\cite{zakharov2019dpod} & PVNet\cite{peng2019pvnet} & CDPN\cite{li2019cdpn} \\
\midrule
Ape   & 4.0 & 11.9 & 35.1 & \textbf{38.9} & 21.6 & 53.3 & 43.6 & \textbf{64.4} \\
Bvise & 20.9 & 66.2 & 59.4 & \textbf{75.2}& 81.8 & 95.2 & \textbf{99.9} & 97.8 \\
Cam   & 30.5 & 22.4 & 15.5 & \textbf{36.9}& 36.6 & 90.0 & 86.9 & \textbf{91.7} \\
Can   & 35.9 & 59.8 & 48.8 & \textbf{65.6}& 68.8 & 94.1 & 95.5 & \textbf{95.9} \\
Cat   & 17.9 & 26.9 & 28.1 & \textbf{57.9}& 41.8 & 60.4 & 79.3 & \textbf{83.8} \\
Drill & 24.0 & 44.6 & 59.3 & \textbf{67.0}& 63.5 & \textbf{97.4} & 96.4 & 96.2 \\
Duck  & 4.9 & 8.3 & \textbf{25.6} & 19.6 & 27.2 & 66.0 & 52.6 & \textbf{66.8} \\
Eggbox& 81.0 & 55.7 & 51.2 &\textbf{99.0}& 69.6 & 99.6 & 99.2 & \textbf{99.7} \\
Glue  & 45.5 & 54.6 & 34.6 & \textbf{94.1}& 80.0 & 93.8 & 95.7 & \textbf{99.6} \\
Holep & 17.6 & 15.5 & \textbf{17.7} & 16.2 & 42.6 & 64.9 & 81.9 & \textbf{85.8} \\
Iron  & 32.0 & 60.8 & \textbf{84.7} & 77.9 & 75.0 & \textbf{99.8} & 98.9 & 97.9 \\
Lamp  & 60.5 & -- & 45.0 &\textbf{68.2} & 71.1 & 88.1 & \textbf{99.3} & 97.9 \\
Phone & 33.8 & 34.4 & 20.9 & \textbf{50.1} & 47.7 & 71.4 & \textbf{92.4} & 90.8 \\
\midrule
Mean  & 31.4 & 38.8 & 40.5 & \textbf{58.9} & 56.0 & 82.6 &  86.3 & \textbf{89.9}\\
\bottomrule
\end{tabular}
}
% \vspace{-0.2mm}
\caption[]{\textbf{Results for \emph{LineMOD}.} \textit{Top}: Qualitative results. We depict the 6D pose by overlaying the 3D bounding box onto the input image. \emph{Blue} constitutes the ground truth pose, \emph{red} and \emph{green} show our results before and after self-supervision. None of the samples has been seen before. \textit{Bottom}: Comparison with state-of-the-art. We present the results for the Average Recall(\%) of ADD(-S) metric. \emph{Real Pose Labels} refers to the 15\% training split from~\cite{Brachmann2014Learning6O} with pose labels. We use the same split for training, however, without employing labels.\footnotemark[2]
}
% \vspace{-2mm}
\label{tab:lm_sota}
\end{table}
In line with other works, we distinguish between training with and without real pose labels, \ie~making use of annotated real training data. Despite exploiting real data, we do not employ any pose labels and must, therefore, be classified as the latter. We want to highlight that our model can produce state-of-the-art results for training with and without labels. Referring to Table~\ref{tab:lm_sota}, for training using only synthetic data, $\pselfLB$ reveals an average recall of $40.1\%$, which is deliberately better than AAE~\cite{sundermeyer2018implicit} with $31.4\%$ and on par with MHP~\cite{manhardt2019ambiguity} and DPOD\footnotemark[2]~\cite{zakharov2019dpod}
reporting $38.8\%$ and $40.5\%$. On the other hand, as for training with real pose labels, we are again on par with other recently published methods such as PVNet~\cite{peng2019pvnet} and CDPN~\cite{li2019cdpn} reporting a mean average recall of $86.9\%$. 
Furthermore, our proposed self-supervision $\pself$ achieves an overall average recall of $58.9\%$, which is more than $51\%$ of relative improvement over all state-of-the-art methods using no real pose labels. 
Except for \emph{Holep}, \emph{Duck} and \emph{Iron}, we can report a significant increase. % for 10 out of 13 objects. 
% As explained before, 
Objects with little variation in color and geometry can become difficult to optimize. 
In addition, the 3D mesh of the \emph{Holep} is rather different 
compared with 
the actual perceived object in the real images, which makes our visual alignment less meaningful.
% This complicates learning from these sequences as our visual alignment becomes less meaningful.
\footnotetext[2]{The numbers of \cite{zakharov2019dpod} and \cite{manhardt2018deep} are different as in their paper since they used average precision instead. The authors provided us with their results for average recall.}

\paragraph{\textbf{HomebrewedDB Dataset.}}
\begin{figure}[t!]
\begin{floatrow}
  \centering
    \scalebox{0.75}{
    \setlength{\tabcolsep}{0.1em}
        \begin{tabular}[b]{@{}c|c|c|c|ccc||c@{}}
        \toprule
                \multirow{2}{*}{Method} & \multicolumn{3}{c|}{Supervision} &\multicolumn{3}{c}{Object} & \\
                \cmidrule{2-8}
                 & Syn & Self & Real GT & Bvise & Driller & Phone & Mean \\
                \midrule
                DPOD~\cite{zakharov2019dpod} & \checkmark & & & 52.9 & 37.8 & 7.3 & 32.7 \\
                SSD6D+Ref. \cite{manhardt2018deep} &  \checkmark & & & \textbf{82.0} & 22.9 & 24.9 & 43.3 \\
                \midrule
                $\pselfLB$ & \checkmark & &  & 37.7 & 19.2 & 20.9 & 25.9 \\
                $\pself$ &  \checkmark & \checkmark & & 72.1 & \textbf{65.1} & \textbf{41.8} & \textbf{59.7} \\
                $\pselfUB$ & \checkmark & & \checkmark  & 71.0 & 64.2 & 46.5 & 60.6 \\ 
                \bottomrule
        \end{tabular}
    }\hspace{-2.5mm}
\ffigbox{%
\centering
  \includegraphics[width=0.86\linewidth,valign=m]{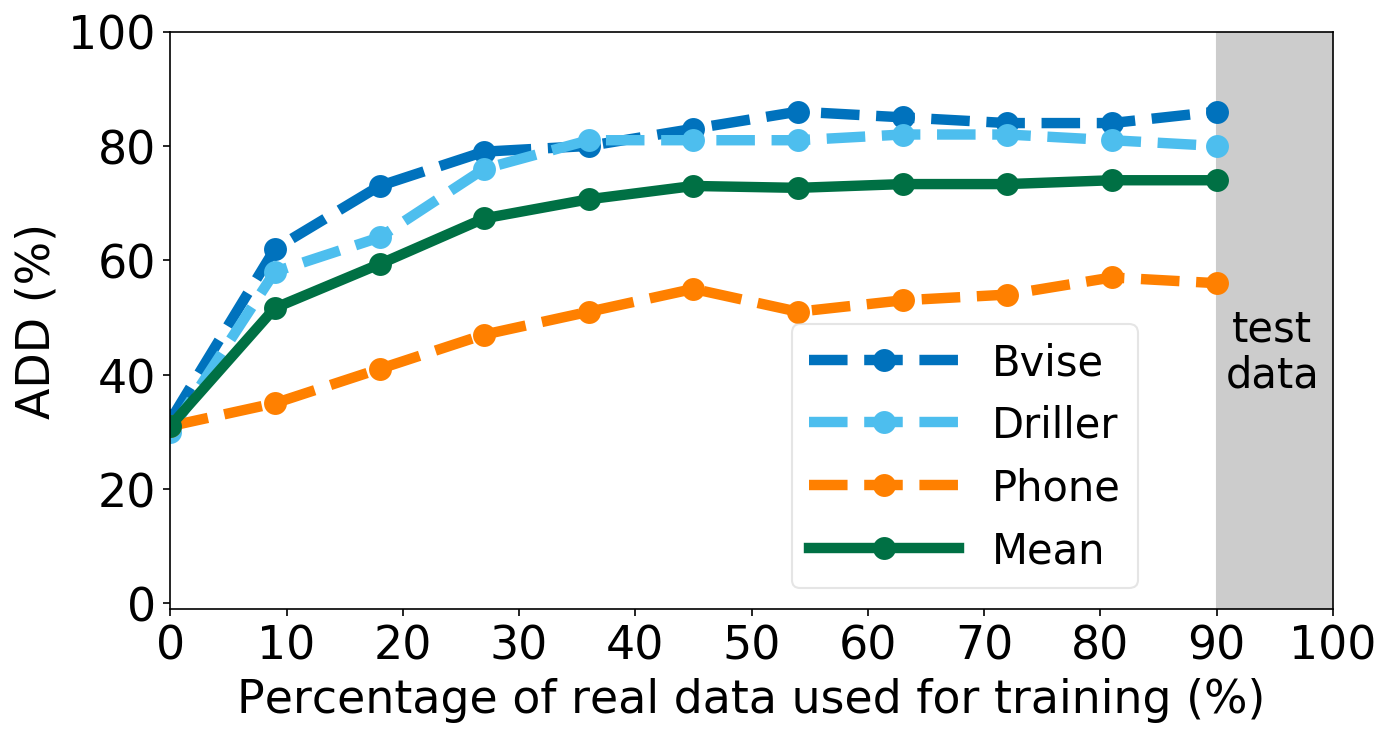}%
}{}
\end{floatrow}
% \vspace{-0.2mm}
\caption[]{\textbf{Results for \emph{HomebrewedDB}.} \textit{Left}: Comparison with~\cite{zakharov2019dpod} and~\cite{manhardt2018deep}.\footnotemark[2] 
We report our results for synthetic data ($\pselfLB$) 
and after self-supervision ($\pself$) or full-supervision ($\pselfUB$) using 15\% of real data from~\cite{kaskman2019homebreweddb}. \textit{Right}: Self-supervised training \wrt an increasing percentage of real training data. Results are always reported on the same unseen test split.}
% \vspace{-2mm}
\label{fig:hb}
\end{figure}
In Fig.~\ref{fig:hb} (\textit{left}) we compare our method with DPOD~\cite{zakharov2019dpod} and SSD6D~\cite{kehl2017ssd} after refinement using~\cite{manhardt2018deep} (SSD6D+Ref.) on three objects of \emph{HomebrewedDB}, which it shares with \emph{LineMOD}.\footnotemark[2] 
Unfortunately, methods directly solving for the 6D pose always implicitly learn the camera intrinsics which degrades the performance when exposed to a new camera. 
2D-3D correspondences based approaches are instead robust to camera changes as they simply run P$n$P using the new intrinsics. Therefore, the performance of our 
$\pselfLB$
% method without self-supervision ($\pselfLB$) 
is slightly outperformed by~\cite{zakharov2019dpod}. 
SSD6D+Ref. ~\cite{manhardt2018deep} employs contour-based pose refinement using renderings for the current hypotheses. 
Similarly, rendering the pose with the new intrinsics enables again easy adaptation and can even exceed \cite{zakharov2019dpod} and our $\pself$ on the \emph{Bvise} object. 
Nevertheless, we can easily adapt to the new domain and intrinsics by only leveraging 15\% of unannotated data from \cite{kaskman2019homebreweddb}. 
In fact, we almost double their numbers for all other objects and reach a similar level as for \emph{LineMOD}. 

Based on this observation, we were curious to understand the adaptation capabilities of our model \wrt the amount of real data that we expose it to.  
We divided the samples from \emph{HomebrewedDB} into 100 images for testing and 900 images for training. 
Afterwards, we repeatedly trained our model with increasing amount of data, however, always evaluating on the same test split. 
In Fig.~\ref{fig:hb} (\textit{right}) we illustrate the corresponding results. 
When using only 15\% (150 samples) of the real data for training, we can already almost double the mean average recall (mAR). 
Using $\approx$~40\% of the real data, the mAR can be improved by $\approx$~130\% from 31\% to 71\%. 
Afterwards, it slowly saturates at $\approx$~74\%.

\paragraph{\textbf{LineMOD Occlusion Dataset.}}
% tried training on other sequences, most are negative, few of them can improve a little (the z prediction relies on the bbox positions a lot)
\begin{table}[t!]
\centering

\scalebox{0.88}{
\setlength{\tabcolsep}{0.3em}
\begin{tabular}{@{}c|c|c|c|cccccccc||c@{}}
\toprule  
\multirow{2}{*}{Method} & \multicolumn{3}{c|}{Supervision} & \multicolumn{9}{c}{Object} \\
\cmidrule{2-13}
& Syn & Self  &  Real GT & Ape  & Can & Cat & Driller & Duck  & Eggbox & Glue  & Holep & Mean \\
\midrule
DPOD~\cite{zakharov2019dpod}  & \checkmark &  &  &  2.3 & 4.0 & 1.2 & 10.5 & 7.2  & 4.4 &  12.9 & 7.5 & 6.3 \\
CDPN~\cite{li2019cdpn}        & \checkmark &  &  & \textbf{20.0} & 15.1 & 16.4 & 5.0 & \textbf{22.2} & 36.1 & 27.9 & \textbf{24.0} & 20.8 \\
\midrule
$\pselfLB$     & \checkmark & & & 7.4 & 14.1 & 7.6 & 18.0 & 12.2 & 18.3 & 31.4 & 11.5 & 15.1 \\
$\pself$ & \checkmark & \checkmark &  & 13.7 & \textbf{43.2} & \textbf{18.7} & \textbf{32.5} & 14.4 & \textbf{57.8} & \textbf{54.3} & 22.0 & \textbf{32.1} \\
$\pselfUB$ & \checkmark &   &   \checkmark  &  47.4 & 79.4 & 56.1 & 83.5 &	48.9 &	90.0 & 93.6 & 62.5 &	70.2 \\
\bottomrule
\end{tabular}
}
\vspace{-0.2mm}
\caption[]{\textbf{Results for \emph{LineMOD Occlusion}.} Comparison with~\cite{zakharov2019dpod} and~\cite{li2019cdpn}. We evaluate the Average Recall(\%) of ADD(-S) on the BOP~\cite{hodan2018bop} split.\footnotemark[3]
}
\vspace{-2mm}
\label{tab:lmo_pose}
\end{table}

\footnotetext[3]{The authors of \cite{zakharov2019dpod} and \cite{li2019cdpn} shared their results for the BOP 2019 challenge~\cite{hodan2018bop}.}
We also evaluate our method on \emph{LineMOD Occlusion} 
% to demonstrate %that $\pself$ can also deal with 
which exhibits
stronger occlusion. 
We follow the BOP~\cite{hodan2018bop} standard and evaluate on a subset of 200 samples. 
We compare $\pself$ with two state-of-the-art methods using synthetic data only, namely DPOD~\cite{zakharov2019dpod} and CDPN~\cite{li2019cdpn}.\footnotemark[3]
While our 
% model before self-supervision
$\pselfLB$
can clearly outperform~\cite{zakharov2019dpod} with 15.1\% compared to 6.3\%, \cite{li2019cdpn} exceeds our $\pselfLB$ by 5.4\% and reports a mean average recall of 20.8\%. 
2D-3D correspondences based methods are more robust towards occlusion as they consider only the visible regions, while direct methods are less stable due to inferring poses from both visible and occluded regions. 
Nonetheless, after utilizing %the omitted samples, 
the remaining real RGB-D data via our self-supervision, 
we can easily surpass~\cite{li2019cdpn} (32.1\% \vs 20.8\%),%. 
% Moreover, we double our original numbers after training with synthetic data. 
~and double the performance of our $\pselfLB$. 
Noteworthy, %as before 
there is still plenty of room for all the methods trained without real labels, compared to our fully-supervised model $\pselfUB$ (70.2\%).

\subsubsection{Domain Adaptation for Pose Estimation\\}

% \begin{table}[t!]
% \centering
% \caption{Mean angle error on Cropped Linemod.}
% \scalebox{0.95}{
% \begin{tabular}{@{}l|c@{}}
% \hline
%           Method & Mean Angle Error \\
% \hline
% PixelDA \cite{bousmalis2017unsupervisedPixelda} & 23.5\degree \\
% \hline
% Syn & 19.8\degree \\
% \hline
% Ours & 17.7\degree \\
% \hline
% \end{tabular}
% }\label{tab:pixelda}
% \end{table}

% transposed table
\begin{table}[t!]
\centering
\scalebox{0.90}{
\setlength{\tabcolsep}{0.3em}
\begin{tabular}{@{}l|c|c|c|c|c@{}}
\toprule
Method & PixelDA \cite{bousmalis2017unsupervisedPixelda} & DRIT~\cite{lee2018diverse} & DeceptionNet~\cite{zakharov2019deceptionnet} & $\pselfLB$ & $\pself$ \\
\midrule
Classification Accuracy (\%) & 99.9 & 98.1 & 95.8 & 100.0 & 100.0 \\
Mean Angle Error (\degree) & 23.5 & 34.4 & 51.9 & 19.8 & \textbf{15.8} \\
\bottomrule
\end{tabular}
}
% \vspace{-0.2mm}
\caption{\textbf{Comparison with state of the art on \emph{Cropped LineMOD}.} We present the classification accuracy as well as mean angle error.}
% \vspace{-2mm}
\label{tab:pixelda}
\end{table}
Since our method is suitable for conducting synthetic to real domain adaptation, we assess transfer skills referring to the commonly used \emph{Cropped LineMOD} scenario. We %essentially 
self-supervise the model with the real training set from \emph{Cropped LineMOD}, and report the mean angle error on the real test set. As shown in Table \ref{tab:pixelda}, our synthetically trained model ($\pselfLB$) slightly exceeds state-of-the-art methods as PixelDA~\cite{bousmalis2017unsupervisedPixelda}. % due to a more advanced network. 
$\pself$ can successfully surpass the original model on the target domain, reducing the mean angle error from $19.8\degree$ to $15.8\degree$.

\section{Conclusion}

This work introduced $\pself$, the first %fully
self-supervised 6D object pose estimation approach aimed at learning from real data without the need for 6D pose annotations. Leveraging neural rendering, we are able to enforce several visual and geometrical constraints, resulting in a remarkable leap forward compared to other state-of-the-art methods. Moreover, $\pself$ demonstrated to notably reduce the gap with the state of the art for pose estimation with real pose labels.%, even surpassing some recently published works such as \cite{rad2017bb8} and \cite{tekin18_yolo6d}. 

A main future direction is exploring how to overcome the need for depth data during self-supervision. Another interesting aspect is to incorporate also 2D detections into self-supervision, as this allows backpropagating the loss in an end-to-end fashion throughout the entire network.

% \section*{Acknowledgements}
\subsubsection{Acknowledgements} 
This work was supported by China Scholarship Council (CSC) Grant \#201906210393.
This work was also supported by the National Key R\&D Program of China under Grant 2018AAA0102801. % ----------------------------------------------------------

% ---- Bibliography --------------------------------------------------
\bibliographystyle{splncs04}
\bibliography{egbib}

\clearpage
\appendix
%%%%%%%%%%%%%%%%%%%%%%%%%%%%%%%%%%%%%%%%%%%%%%%%%%%%%%%%%%%%%%%%%%%%%%%%%%%%%%%%%%%%%%%%%%%%%%%%%%%%%%%%%%%%%
\section{Implementation Details}

\subsection{More Architecture Details}
As mentioned in Section~\ref{sec:method} of the main paper, we employ different lightweight branches to predict the following outputs from the fused FPN feature maps: the 3D rotation $R$ parameterized as a 4D quaternion $q$, the 3D translation $t$ constituted by the 2D projection $(c_x, c_y)$ of the 3D object centroid together with the distance $z$, and the visible object mask $M^P$.

For fusion of the FPN feature maps, we first apply a convolutional layer to reduce the dimension of each layer from 128 to 64, we then conduct bilinearly rescaling for each layer to make them spatially equal with the largest feature map (\ie, $60\times 80$). Finally, we concatenate all different levels to obtain the fused FPN feature map.

Each of our lightweight branches consists of 4 convolutional layers (except 2 for the 2D centroid) with Group Normalization~\citeA{wu2018groupA} and Leaky ReLU~\citeA{xu2015empiricalA} having a negative slope of $0.1$. 
The final output of each predictor is obtained by employing either a $3\times 3$ convolutional layer (\emph{mask}) or a fully connected layer employed after flattening the output of forelast layer (\emph{centroid}, \emph{distance} and \emph{rotation}).

In line with other works~\citeA{zakharov2019dpodA,manhardt2019roiA}, we freeze the parameters in the first 3 stages of the backbone, to reduce the risk of overfitting to synthetic data. We additionally apply various augmentations to the training RGB images, similar to \citeA{sundermeyer2018implicitA,kehl2017ssdA}.

%%%%%%%%%%%%%%%%%%%%%%%%%%%%%%%%%%%%%%%%%%%%%%%%%%%%%%%%%%
\subsection{Experimental Setup} % ========================
We implemented our method in PyTorch 1.3~\citeA{paszke2019pytorchA} and ran all our experiments on a NVIDIA TitanX GPU. In the first stage, we trained our model with a batch size of 12 for 8 epochs. 
During self-supervision, we trained the model for another 100 epochs with a batch size of 3. We utilized RAdam~\citeA{liu2019radamA} combined with Lookahead~\citeA{zhang2019lookahead} for optimization. 
The initial learning rate was set to $10^{-4}$ and decayed after $72\%$ of the training phase using a cosine schedule~\citeA{loshchilov-ICLR17SGDRA}. 
\subsubsection{Employed Hyper-Parameters for Losses.}
Table~\ref{tab:hyper_syn} shows the employed hyper-parameters for training with synthetic data 
\begin{equation}
\loss_{synthetic} \coloneqq \lambda_{class} \loss_{focal} + \lambda_{box} \loss_{giou} + \lambda_{mask} \loss_{bce} + \lambda_{pose} \loss_{pose},
\end{equation}
and self-supervised training on real data
\begin{equation}
\loss_{Self} 
= \loss_{visual} + \eta \loss_{geom},
\end{equation}
\begin{center}
    with
\end{center}
\begin{equation}
     \loss_{visual} = \loss_{mask} + \alpha  \loss_{ab} + \beta \loss_{ms\text{-}ssim} + \gamma \loss_{perceptual}.
\end{equation}
The hyper-parameters are empirically chosen to level the different loss contributions.

\begin{table}[H]
\centering
\scalebox{0.99}{
\setlength{\tabcolsep}{0.3em}
\begin{tabular}{@{}c|cccc|cccc@{}}
\toprule
hyper-parameter & $\lambda_{class}$ & $\lambda_{box}$ & $\lambda_{mask}$ & $\lambda_{pose}$ & $\alpha$ & $\beta$ & $\gamma$ & $\eta$ \\
\midrule
value & 1.0 & 1.0 & 1.0 & 10.0 & 0.2 & 1.0 & 0.15 & 100 \\
\bottomrule
\end{tabular}
}
% \vspace{-0.1mm}
\caption{Employed hyper-parameters for losses.}
% \vspace{-1mm}
\label{tab:hyper_syn}
\end{table}
%%%%%%%%%%%%%%%%%%%%%%%%%%%%%%%%%%%%%%%%%%%%%%%%%%%%%%%%%%
\subsection{Details of $\loss_{ms\text{-}ssim}$}
In this section, we present the details of MS-SSIM loss ($\loss_{ms\text{-}ssim}$), which is an important part for the visual alignment ($\loss_{visual}$) referring to our proposed self-supervision ($\loss_{Self}$). 
The structural similarity index (SSIM) ~\citeA{wang2004ssimA} for pixel $p$ is defined as
\begin{equation}\label{eg:ssim}
ssim(p) =
\frac{2\mu_x \mu_y + c_1}{\mu_x^2 + \mu_y^2 + c_1}
    \cdot 
    \frac{2\sigma_{xy} + c_2}{\sigma_x^2 + \sigma_y^2 + c_2} =  l(p)\cdot cs(p).
\end{equation}
Thereby, SSIM is computed on blocks, with $(\mu_x, \mu_y)$ and 
$\big(\begin{smallmatrix}
  \sigma_x^2 & \sigma_{xy}\\
  \sigma_{xy} & \sigma_y^2
\end{smallmatrix}\big)$ 
denoting the corresponding means and covariance matrix \wrt $p$, respectively. 
In practice, we employ two constants $c_1=0.01$ and $c_2=0.03$ for numerical stability.
In \citeA{zhao2016lossA}, the authors extend this measurement to a perceptually-motivated loss function referring to a widely used multi-scale version of SSIM, namely MS-SSIM. 
Given a dyadic pyramid of $s$ levels, MS-SSIM can be written as
\begin{equation}
\label{eg:msssim}
ms\text{-}ssim(p) =l_s^\alpha (p) \cdot \prod_{j=1}^s cs_j^{\beta_j} (p),
\end{equation}
where $l_s$ and $cs_j$ are the functions defined on the right in Eq. (\ref{eg:ssim}) at scale $s$ and $j$, respectively.
As in \citeA{zhao2016lossA}, we set $\alpha=1$, $\beta_j=1$, $\forall j \in \{1, ..., s\}$, and $s=5$.
For convenience, given two RGB images $I_A$ and $I_B$, we denote the MS-SSIM between them as $ms\text{-}ssim(I_A, I_B, s)$.  Since, $ms\text{-}ssim$ measure the similarity between two samples with 1 being the maximum, we rewrite the MS-SSIM loss as 
%------------------------
\begin{equation}
\label{eg:loss_msssim}
\loss_{ms\text{-}ssim} \coloneqq
1 - ms\text{-}ssim(I_A, I_B, s).
\end{equation}
% \newpage

%%%%%%%%%%%%%%%%%%%%%%%%%%%%%%%%%%%%%%%%%%%%%%%%%%%%%%%%%%%
\section{Detailed Analysis on the Quality of Predicted Masks}

\begin{figure}[H]
	\centering
	\includegraphics[width = 0.99\linewidth]{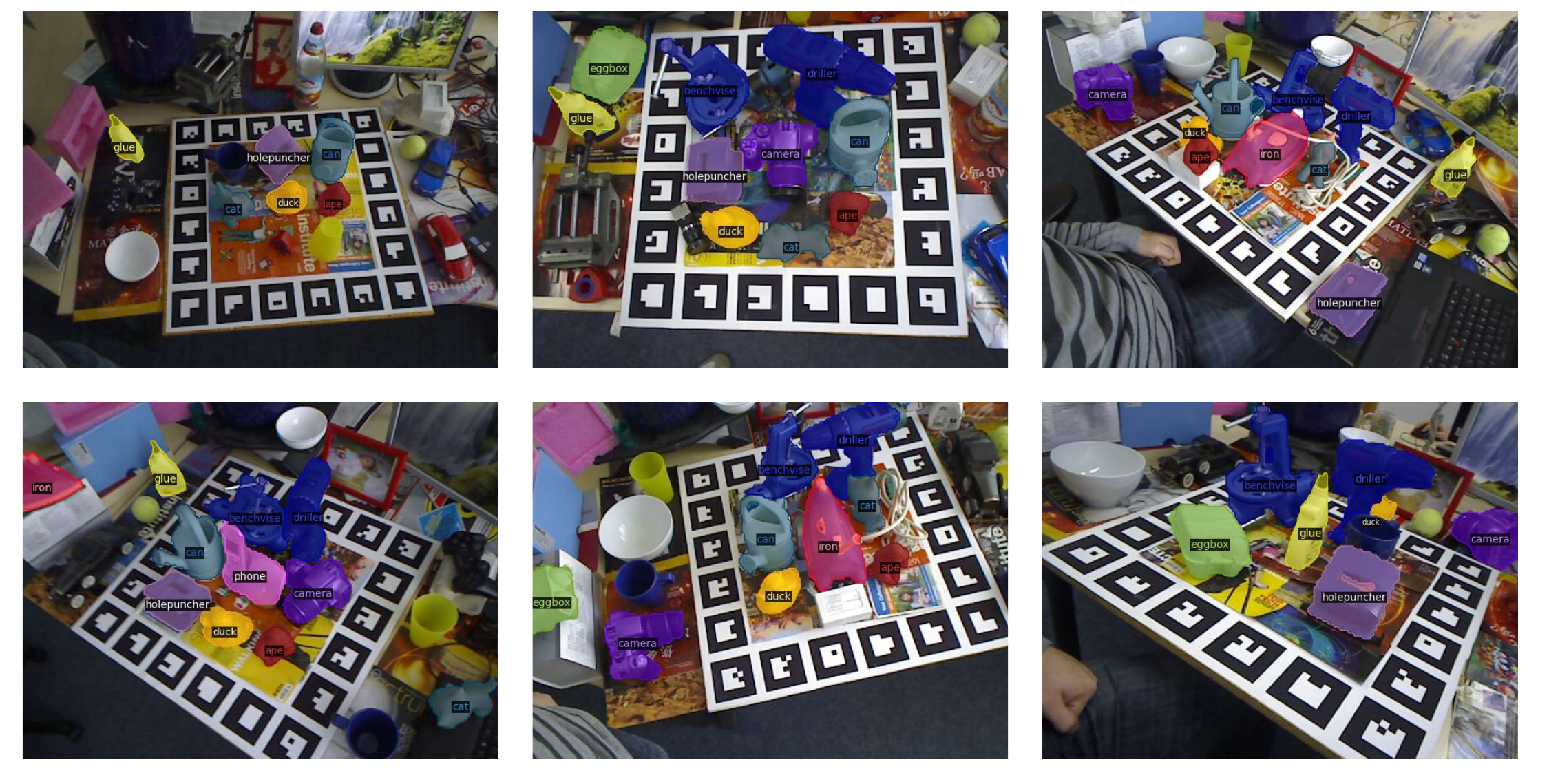}
	\caption{Qualitative results for predicted masks on \emph{LineMOD} test set. (Best viewed zoomed-in, in color).
	}
	\label{fig:lm_mask_vis}
\end{figure}

\begin{table}[H]
\centering

\scalebox{0.83}{
\setlength{\tabcolsep}{0.2em}
\begin{tabular}{@{}l|c|c|c|c|c|c|c|c|c|c|c|c|c||c@{}}
\toprule
& Ape   & Bvise & Cam & Can & Cat & Drill & Duck  & Eggbox & Glue  & Holep & Iron  & Lamp  & Phone & Mean \\
\midrule
F1 score  & 94.44	&92.04	&81.57&	92.51&	91.60&	91.51	&92.17&	94.82&	89.57&	91.02&	91.31&	78.16&	84.52&	89.63 \\
\midrule
IoU & 94.72 &	92.45 &	83.41 &	92.88 &	92.08 &	92.03 &	93.76 &	95.02 &	90.35 &	91.90 &	91.80 &	79.13 &	85.39 &	90.38 \\
\bottomrule
\end{tabular}
}
\caption{\textbf{Detailed evaluation of predicted masks.} We report the F1 score(\%) and IoU(\%) on \emph{LineMOD} test set.}
\label{tab:lm_mask}
\end{table}
Since our self-supervision relies on high quality of the predicted masks from our synthetically trained model $\pselfLB$, we present the detailed quantitative evaluation in Table~\ref{tab:lm_mask}. Specifically, we calculated F1 score and IoU between the predicted masks and the ground-truth masks on \emph{LineMOD} test set. We can see that the predicted masks are very accurate on almost all objects, with the average F1 score 89.63\% and mIoU 90.38\%.
Thus we can regard the predicted masks from $\pselfLB$ as a reliable self-supervision signal. Fig.~\ref{fig:lm_mask_vis} shows some qualitative examples for the predicted masks on \emph{LineMOD} test set.

%%%%%%%%%%%%%%%%%%%%%%%%%%%%%%%%%%%%%%%%%%%%%%%%%%%%%%%
\section{More Qualitative Results}  % ====================
In this section, we want to present additional qualitative results for each conducted experiment. 

\begin{figure}[H]
	\centering
	\includegraphics[width = 0.99\linewidth]{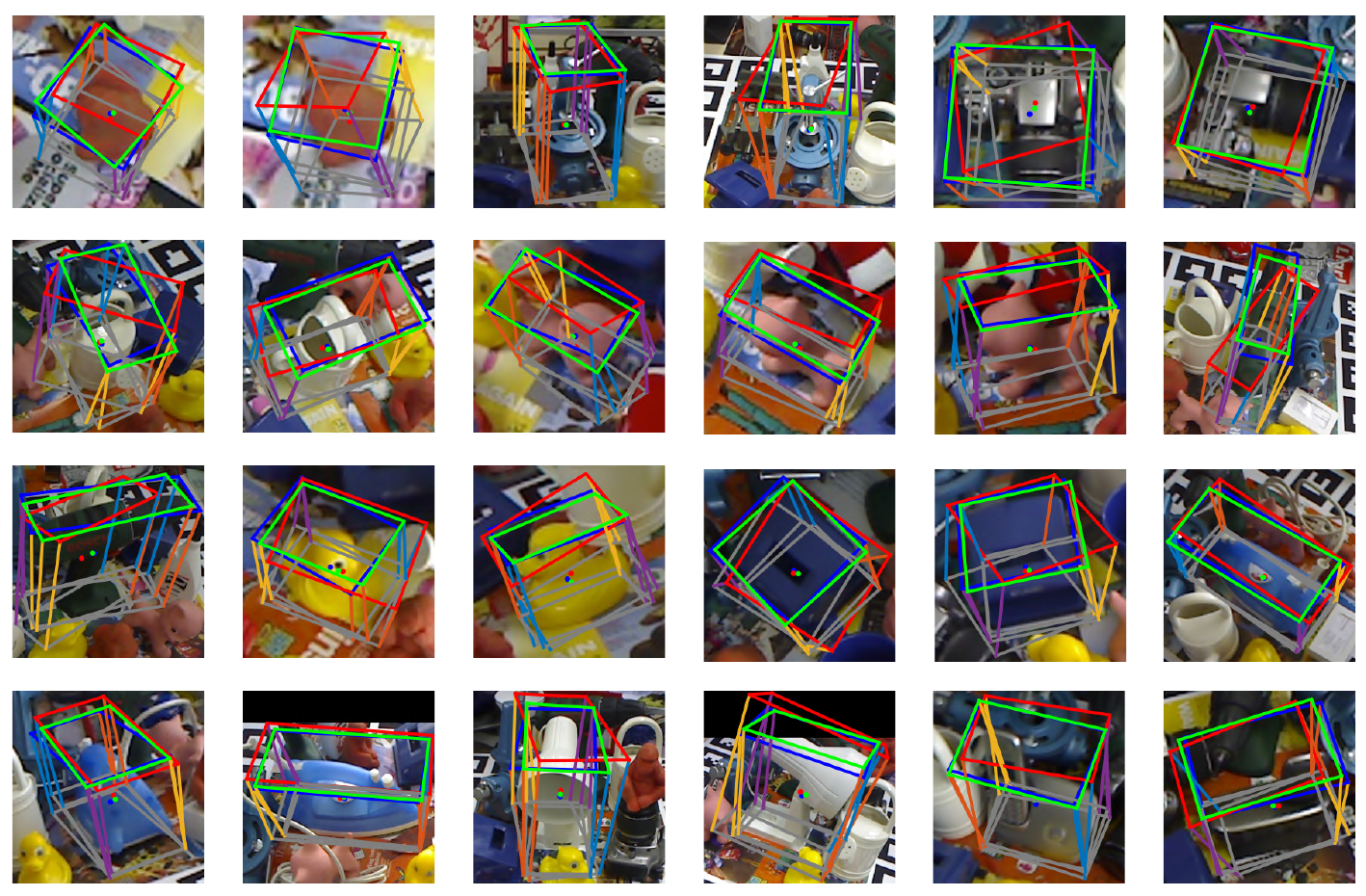}
	\caption{Qualitative results on \emph{Cropped LineMOD}.
	We visualize the 6D pose by overlaying the image with the corresponding transformed 3D bounding box.
	While \textit{Blue} constitutes the ground truth pose, we demonstrate in \textit{Red} and \textit{Green} the results before and after applying our self-supervision, respectively.
	}
	\label{fig:lm_crop_qualitative}
\end{figure}

% \smallskip
\noindent \textbf{Domain Adaption.} In Fig.~\ref{fig:lm_crop_qualitative}, we demonstrate our results for domain adaption, leveraging the \emph{Cropped LineMOD} dataset~\citeA{wohlhart2015learningA}. While we constitute the ground truth poses in \emph{Blue}, we demonstrate in \emph{Red} and \emph{Green} the results before (Self-LB) and after applying our self-supervision ($\pself$), respectively.

\begin{figure}[H]
	\centering
	\includegraphics[width = 0.99\linewidth]{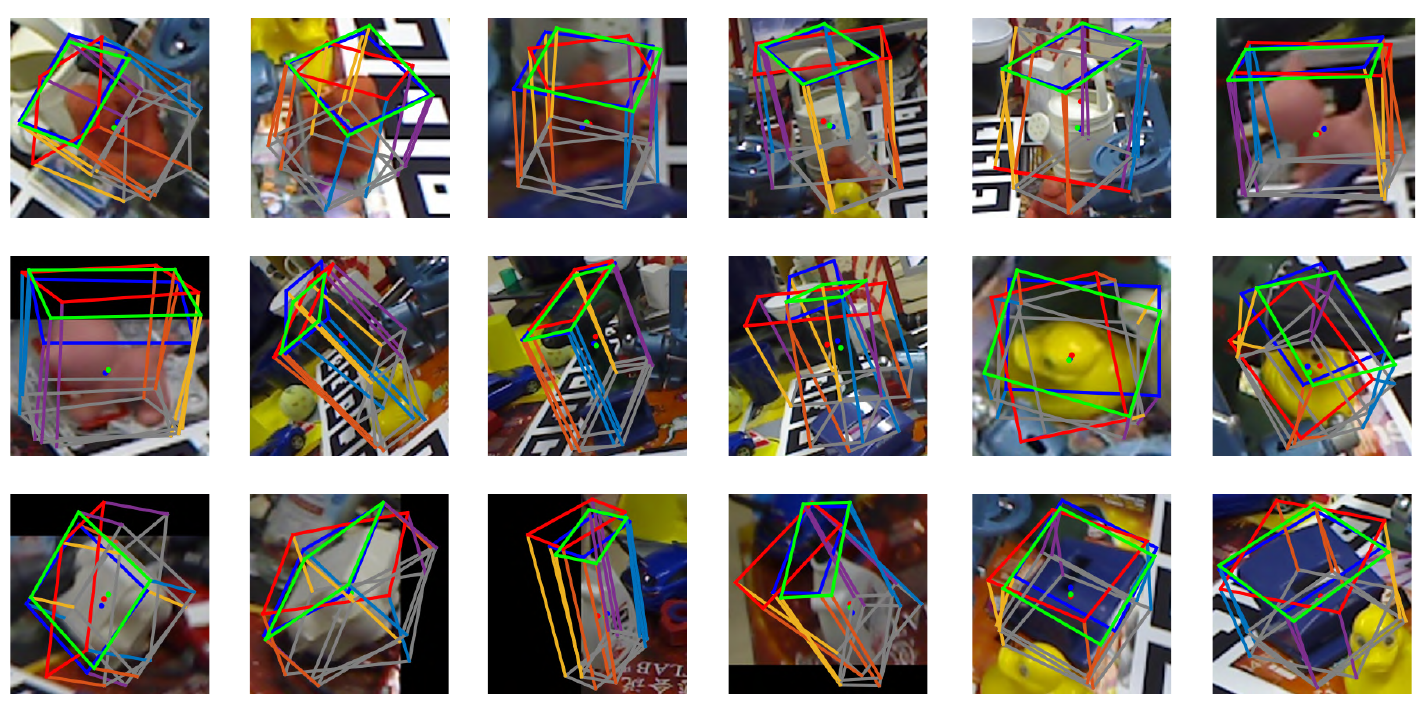}
	\caption{Qualitative results on \emph{LineMOD Occlusion}.
	We visualize the 6D pose by overlaying the image with the corresponding transformed 3D bounding box.
	While \textit{Blue} constitutes the ground truth pose, we demonstrate in \textit{Red} and \textit{Green} the results before and after applying our self-supervision, respectively.
	}
	\label{fig:lmo_qualitative}
\end{figure}

\begin{figure}[H]
	\centering
	\includegraphics[width = 0.99\linewidth]{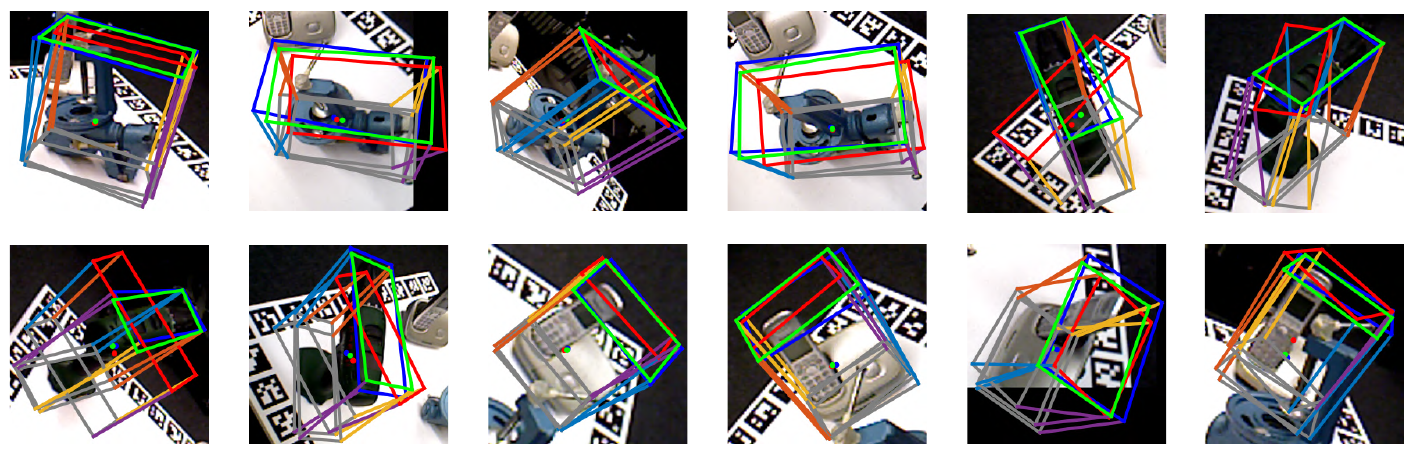}
% 	\vspace{-0.2mm}
	\caption{Qualitative results on \emph{HomebrewedDB}.
	We visualize the 6D pose by overlaying the image with the corresponding transformed 3D bounding box.
	While \textit{Blue} constitutes the ground truth pose, we demonstrate in \textit{Red} and \textit{Green} the results before and after applying our self-supervision, respectively.
	}
% 	\vspace{-3mm}
	\label{fig:hb_qualitative}
\end{figure}

\begin{figure}[H]
	\centering
	\includegraphics[width = 0.99\linewidth]{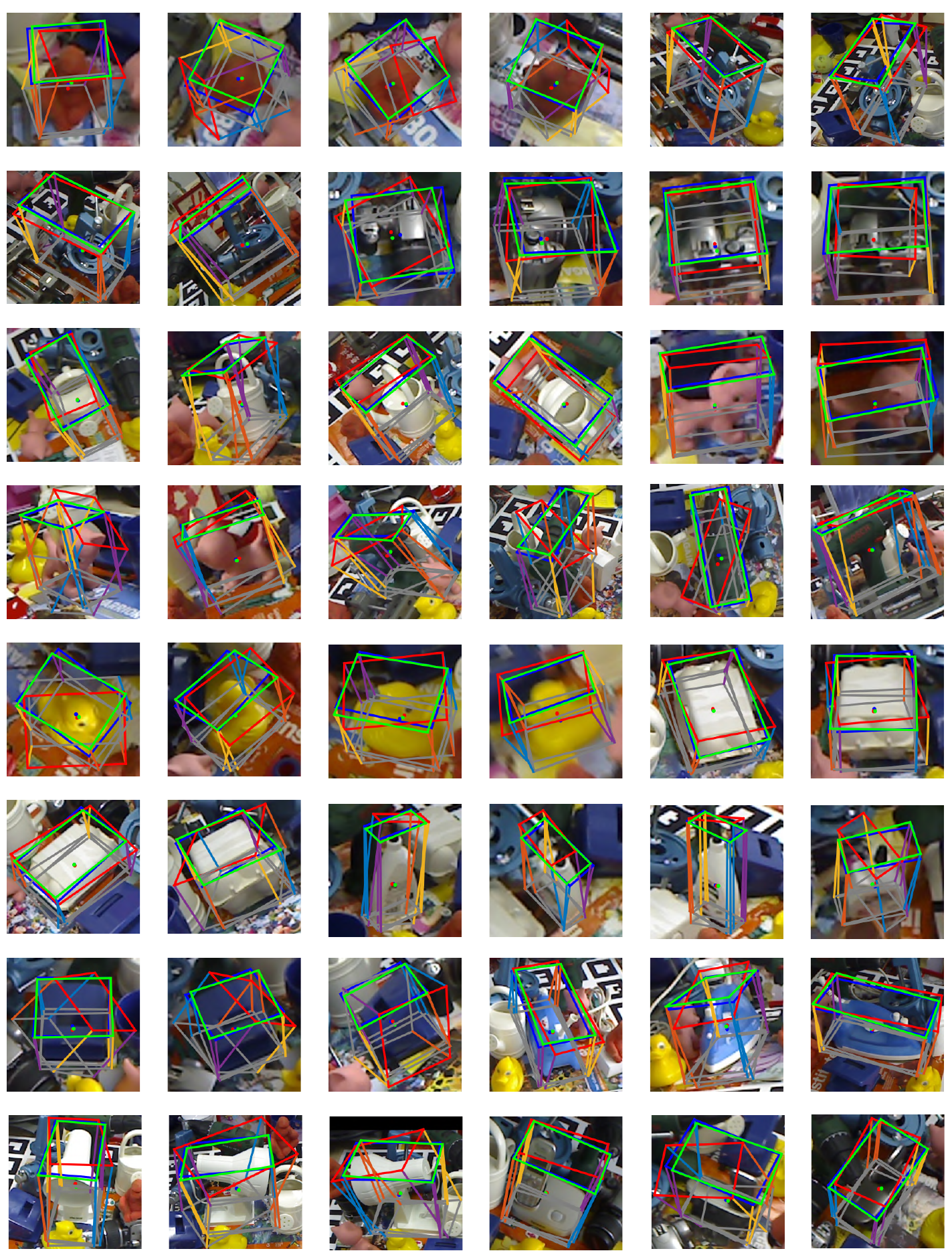}
	\caption{More qualitative results on \emph{LineMOD}.
	We visualize the 6D pose by overlaying the image with the corresponding transformed 3D bounding box.
	While \textit{Blue} constitutes the ground truth pose, we demonstrate in \textit{Red} and \textit{Green} the results before and after applying our self-supervision, respectively.
	}
	\label{fig:lm_qualitative_more}
\end{figure}

% \newpage
\noindent \textbf{6D Pose Estimation}
We additionally present qualitative results for 6D pose estimation. Notice that we again depict the ground truth pose in \emph{Blue} and the prediction before ($\pselfLB$) and after self-supervision ($\pself$) in \emph{Red} and \emph{Green}. While the initial results are very noisy in terms of 6D pose, the self-supervised model produces highly accurate pose estimates \wrt the given ground truth.
\newpage
% \smallskip
\noindent The results on LineMOD Occlusion~\citeA{Brachmann2014Learning6OA}, HomebrewedDB~\citeA{kaskman2019homebreweddbA}, and LineMOD~\citeA{Hinterstoisser2012A} are respectively depicted in Fig.~\ref{fig:lmo_qualitative}, Fig.~\ref{fig:hb_qualitative}, and Fig.~\ref{fig:lm_qualitative_more}.

\newpage
\newpage

\begin{figure}[H]
	\centering
	\includegraphics[width = 0.99\linewidth]{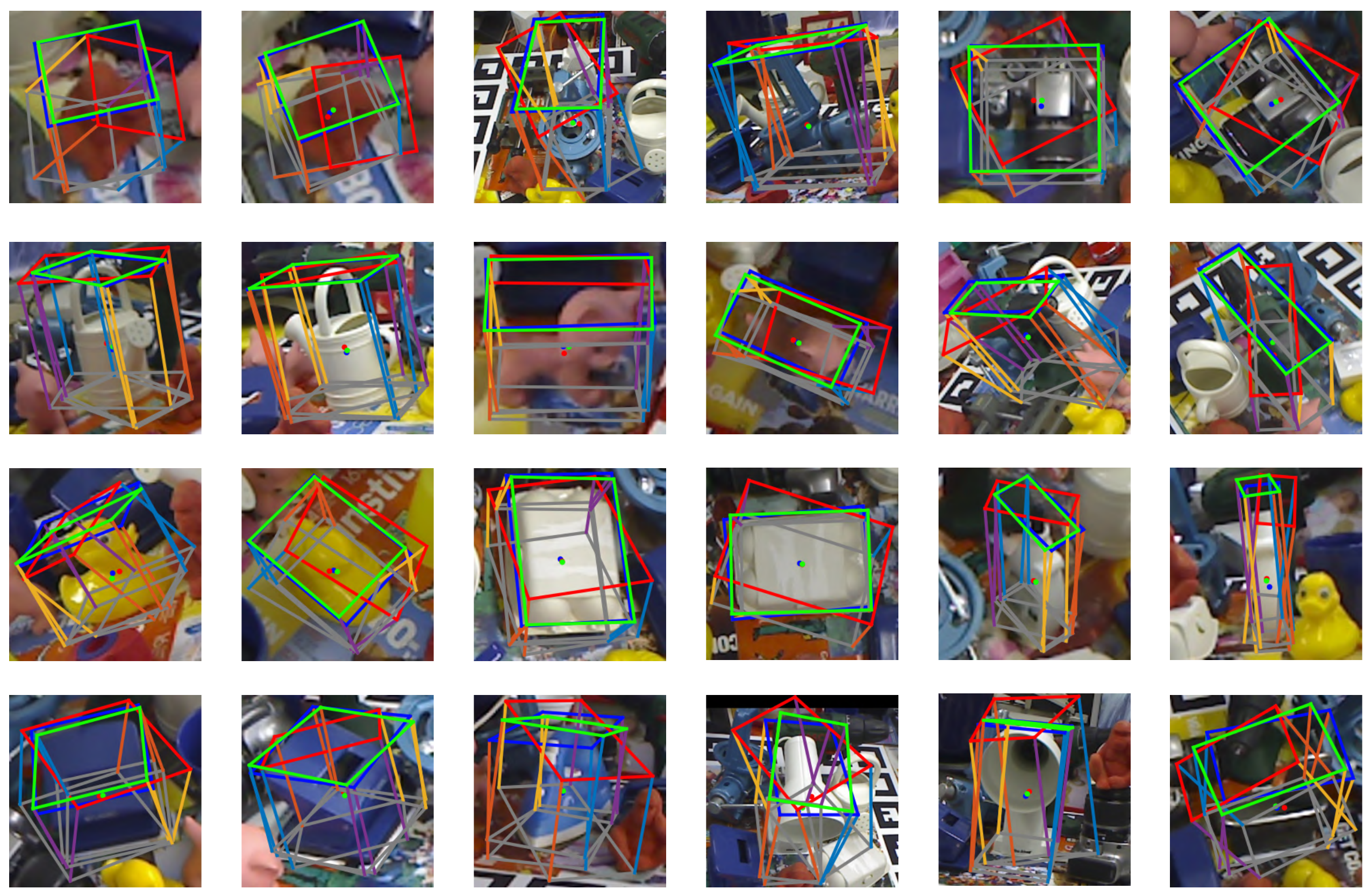}
	\caption{More qualitative results on \textit{Self-Supervision~\vs~6D Pose Error}.
	We visualize the 6D pose by overlaying the image with the corresponding transformed 3D bounding box.
	While \textit{Blue} constitutes the ground truth pose, we demonstrate in \textit{Red} and \textit{Green} the results before and after applying our self-supervision to the single image for 200 iterations, respectively.
	}
	\label{fig:iter_qualitative}
\end{figure}

\noindent \textbf{Self-Supervision~\vs~6D Pose Error.}
In the main paper, we demonstrate the existence of a strong correlation between our self-supervised loss function and the 6D pose accuracy. In particular, we conduct our self-supervision separately to single images for in total 200 iterations. In Fig.~\ref{fig:iter_qualitative}, we illustrate additional qualitative results for this experiment. Despite not possessing any pose labels, we can always compute pose estimates (\emph{Green}) which are almost perfectly aligned with the ground truth (\emph{Blue}). This is even true for poses, which are badly initialized using $\pselfLB$ (\emph{Red}). This strongly suggest that our loss function is able to solve for the 6D pose, while actually optimizing for visual and geometric alignment.
\newpage
%%%%%%%%%%%%%%%%%%%%%%%%%%%%%%%%%%%%%%%%%%%%%%%%%%%%%%%
%%%%%%%%%%%%%%%%%%%%%%%%%%%%%%%%%%%%%%%%%%%%%%%%%%%%%%
\section{Results on YCB-Video}
% ---------------------------
\vspace{-2.5em}
\begin{table}[H]
% \centering
\begin{floatrow}
   \centering
    \scalebox{0.99}{
    \setlength{\tabcolsep}{0.3em}
        \begin{tabular}[b]{@{}l|c|c@{}}
        \toprule
        &$\pselfLB$ & $\pself$    \\
        \midrule
        006\_mustard\_bottle & 73.7 & \textbf{88.2} \\
        007\_tuna\_fish\_can & 26.6 & \textbf{69.7} \\
        011\_banana & 4.0 & \textbf{10.3} \\
        025\_mug & 23.9 & \textbf{43.4} \\
        035\_power\_drill & 21.4 & \textbf{31.4} \\
        \midrule
        Mean & 29.9 & \textbf{48.6} \\
        \bottomrule
        \end{tabular}
    }
\ffigbox{
    \centering
    \includegraphics[width=0.65\linewidth,valign=m]{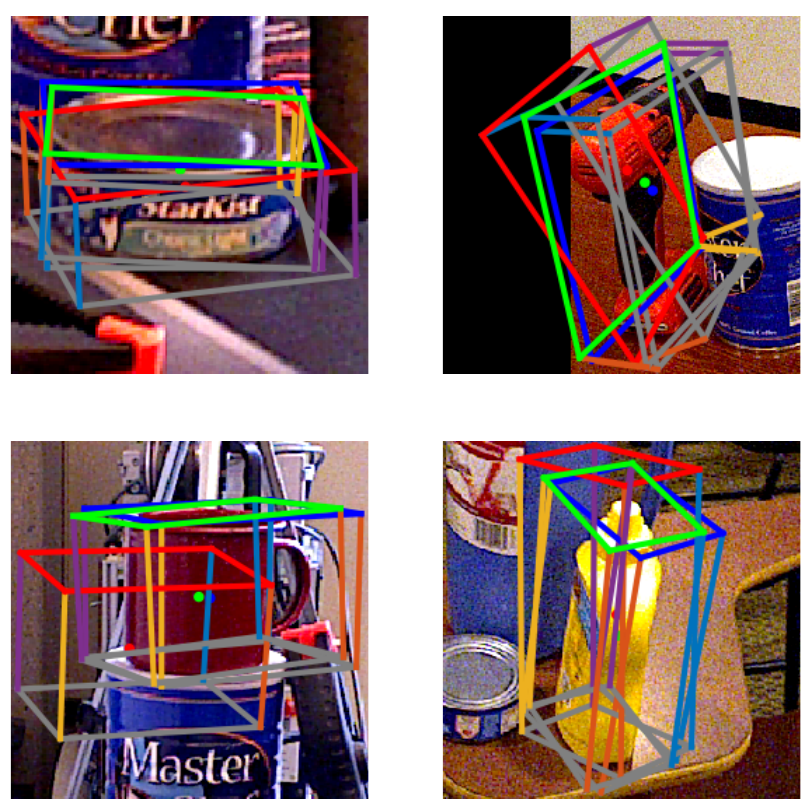}%
}{}
\end{floatrow}
% \vspace{-0.2mm}
\caption[]{\textit{Left}: Results on YCB-Video~\citeA{xiang2017posecnnA} dataset. We report Average Recall (\%) of ADD-S for the top 4 objects and ADD for 035\_power\_drill. 
\textit{Right}: Qualitative results.}
% \vspace{-2mm}
\label{tab:ycb}
\end{table}

Although being out of the scope for the main body of this work, we further evaluated our method on a subset (5 objects) of YCB-Video~\citeA{xiang2017posecnnA}, in order to emphasize that $\pself$ can be applied to any 6D pose scenario.

We first fully supervise $\pselfLB$ on a mixture of synthetic RGB images from YCB-Video and photorealistic renderings from~\citeA{tremblay2018fallingA}.
We then self-supervise the model on 10\% of the real RGB-D images for each of these 5 objects from the training set, however, without leveraging any 6D annotations ($\pself$).
Since the first four objects exhibit rotational symmetries, we report the average recall for ADD-S for them. For the remaining object (\ie~\emph{035\_power\_drill}), we refer to ADD instead, as it does not possess rotational symmetries. In line with all other experiments, we can constitute that leveraging our self-supervision, we are able to significantly enhance the average recall \wrt ADD(-S) for each object (Table~\ref{tab:ycb}).

\bibliographystyleA{splncs04}
\bibliographyA{egbib}
\end{document}